\documentclass[conference]{IEEEtran}
\usepackage{times}
\usepackage{bm}
\usepackage{amsmath}

\usepackage[numbers]{natbib}
\usepackage{multicol}
\usepackage{booktabs}
\usepackage[bookmarks=true]{hyperref}
\usepackage{graphicx}
\usepackage{subcaption}
\usepackage[font=small,labelfont=bf]{caption}
\usepackage{algorithm}
\usepackage{algpseudocode}
\usepackage{amssymb}

\newcommand{\mc}{\mathcal}

\usepackage{xcolor}
\usepackage{colortbl}
\usepackage{etoolbox}
\def\tablename{Table}
\newcommand{\minew}[1]{\textcolor{black}{#1}}

\newcommand{\tabincell}[2]{\begin{tabular}{@{}#1@{}}#2\end{tabular}}

\newcounter{reviewer}
\newcounter{point}[reviewer]
\usepackage{xcolor}
\usepackage{colortbl}

\usepackage{authblk}
\usepackage{tcolorbox}
\IEEEoverridecommandlockouts
\begin{document}

\title{Logic-Skill Programming: An Optimization-based Approach to Sequential Skill Planning}


\author{%
  \makebox{Teng Xue\textsuperscript{1,2}, Amirreza Razmjoo\textsuperscript{1,2}, Suhan Shetty\textsuperscript{1,2}, Sylvain Calinon\textsuperscript{1, 2}} \\
  \makebox{\textsuperscript{1}Idiap Research Institute \; \textsuperscript{2}École Polytechnique Fédérale de Lausanne (EPFL)} 
}

\maketitle

\begin{abstract}
Recent advances in robot skill learning have unlocked the potential to construct task-agnostic skill libraries, facilitating the seamless sequencing of multiple simple manipulation primitives (aka. skills) to tackle significantly more complex tasks. Nevertheless, determining the optimal sequence for independently learned skills remains an open problem, particularly when the objective is given solely in terms of the final geometric configuration rather than a symbolic goal. To address this challenge, we propose Logic-Skill Programming (LSP), an optimization-based approach that sequences independently learned skills to solve long-horizon tasks. We formulate a first-order extension of a mathematical program to optimize the overall cumulative reward of all skills within a plan, abstracted by the sum of value functions. To solve such programs, we leverage the use of tensor train factorization to construct the value function space, and rely on alternations between symbolic search and skill value optimization to find the appropriate skill skeleton and optimal subgoal sequence. Experimental results indicate that the obtained value functions provide a superior approximation of cumulative rewards compared to state-of-the-art reinforcement learning methods. Furthermore, we validate LSP in three manipulation domains, encompassing both prehensile and non-prehensile primitives. The results demonstrate its capability to identify the optimal solution over the full logic and geometric path. The real-robot experiments showcase the effectiveness of our approach to cope with contact uncertainty and external disturbances in the real world. Project webpage: \href{https://sites.google.com/view/lsp4plan}{https://sites.google.com/view/lsp4plan}.

\end{abstract}

\IEEEpeerreviewmaketitle

\section{Introduction}

Consider the following task: "A large box is positioned on the table, next to a wall. The objective is to reorient the box to a new 6D pose using a single robot manipulator with minimal control efforts (or maximal rewards). The robot is allowed to have any interactions with the surroundings." A potential solution involves pushing the box until it reaches the wall, then pivoting against the wall, followed by pulling it to the target. It is noteworthy that the objective is only given in terms of the evaluation of the final geometric configuration, and potential control costs.

\begin{figure}[t]
    \centering
    \includegraphics[width=0.5\textwidth]{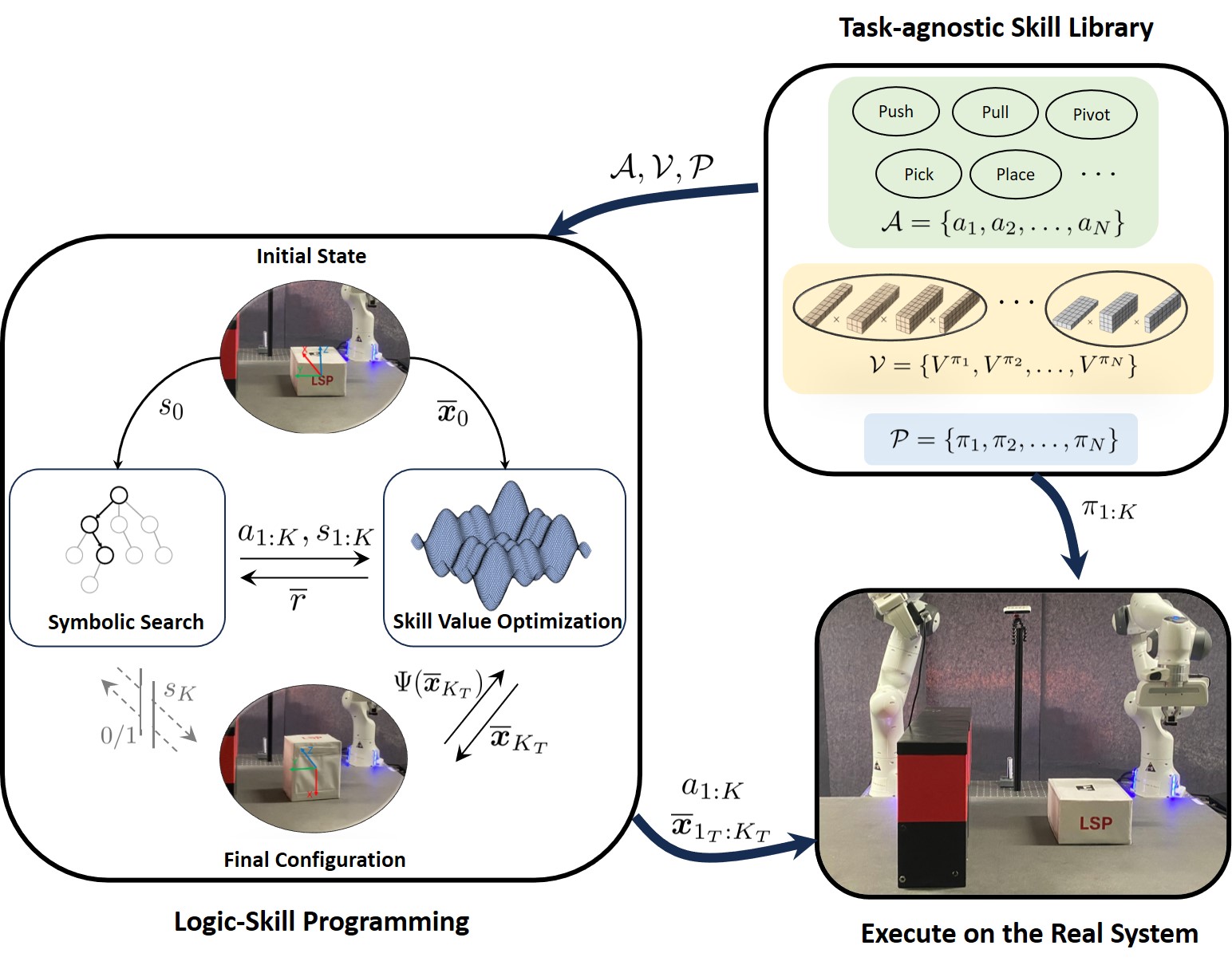}
    \caption{\minew{Overview of the proposed approach: Given the evaluation function $\Psi$ of the final configuration, along with the initial symbolic state $s_0$ and geometric state $\overline{\bm{x}}_0$, the objective of LSP is to find a solution that can accomplish the task with minimal control costs. A task-agnostic skill library is pretrained, consisting of $N$ skill operators $\mathcal{A} = \{a_{1:N}\}$, along with corresponding value functions $\mathcal{V} = \{V^{\pi_{1:N}}\}$ and policies $\mathcal{P}=\{\pi_{1:N}\}$ in Tensor Train format. LSP solves this problem by alternating between symbolic search and skill value optimization for joint logic-geometric reasoning. Symbols $s_{1:K}$ are used as constraints for skill optimization, while skill optimization is used to check skeleton feasibility and final configuration performance, with a feedback reward $\overline{r}$ informing the symbolic search. This results in the appropriate skill skeleton $a_{1:K}$ and subgoal sequence $\overline{\bm{x}}_{1_T:K_T}$, which are then combined with the skill policies $\pi_{1:K}$ existing in the skill library to actuate the real robot. Notably, the gray channel with symbolic final state $s_K$ is interrupted because our framework eliminates the need for a symbolic target goal $s_T$, while such information is typically required in existing sampling-based sequential skill planning methods.}}
    \label{fig:pipeline}
\end{figure}

Such tasks are quite common in sequential manipulation scenarios, typically involving the sequencing of multiple manipulation primitives, such as push, pivot, and pull, to achieve a long-horizon target with sparse rewards. Solving these tasks requires reasoning about the appropriate sequence of primitives and corresponding motion trajectories. This hybrid structure results in combinatorial complexity, making it expensive to find a solution. To address this challenge, Mixed-Integer Programming (MIP) \cite{marcucci2019mixed} is an intuitive approach that does not require a careful design of the system model but relies on branch-and-bound techniques to efficiently prune solutions. While MIP works well with convex optimization formulations, such as graph of convex sets \cite{marcucci2023motion}, manipulation tasks involving interactions with the surroundings usually violate this assumption. Another approach is to implicitly model the hybrid system with complementary constraints \cite{posa2014direct, moura2022non}. By uncovering the internal structures of different manipulation primitives, this method eliminates the need for integer variables in problem formulation, allowing the use of continuous optimization techniques. However, it often leads to poor local optima.


A more general approach involves using logic as a combinatorial expression of possible manipulation primitives \cite{toussaint2018differentiable}. This approach follows the principles of MIP but extends the formulation to the first-order logic level, allowing the utilization of powerful classical AI techniques. This idea is related to a field known as Task and Motion Planning (TAMP) \cite{garrett2021integrated}, where the objective is to find a feasible or optimal plan for a complicated task given full knowledge of the system and environments. Various approaches have been proposed to solve TAMP problems, including PDDLStream \cite{garrett2020pddlstream} and Logic-Geometric Programming \cite{toussaint2015logic}. TAMP methods have demonstrated high performance in diverse scenarios, such as construction assembly \cite{hartmann2022long}, table rearrangement \cite{quintero2023optimal}, and mobile manipulation \cite{yang2022sequence}. However, they often exhibit poor performance in many realistic scenarios involving model uncertainty and external disturbances. For instance, considering the task introduced at the beginning, although we can derive an offline feasible/optimal trajectory using predefined contact parameters like friction coefficients and damping parameters, this trajectory can be totally wrong because we cannot know exactly the contact model between the robot, objects, and the environment.

Thanks to the advances in learning-based techniques, it is now possible to learn a powerful skill policy for each specific manipulation primitive. The policy can be seen as a short-horizon model predictive controller (MPC) with a good terminal cost function, showing robust and reactive behaviors in the face of model uncertainty and external disturbances. When compared with methods that require online planning \cite{xue2023guided, hogan2020reactive}, policy-based methods \cite{sutton2018reinforcement, Shetty24ICLR, chi2023diffusion} usually exhibit good performance in physical manipulation tasks that need to cope with contact uncertainty. 

We can construct a skill library composed of multiple task-agnostic policies trained independently. This allows for the iterative augmentation of the skill library in a lifelong manner. However, sequencing such task-agnostic policies remains an open problem. Two questions should be considered:


1) Within the library, multiple skills are available. To tackle an unseen, long-horizon manipulation task, the questions arise: which skills should be employed, and what is the correct order?

2) The acquired skills are task-agnostic. Each skill depends on a specific subgoal to generate the appropriate action given the current state. How to determine the subgoal sequence given a skill skeleton?

There has been prior work aimed at addressing these questions \cite{agia2023stap, huang2019continuous, wu2021example}. However, most existing approaches necessitate a well-defined task planning problem with an explicit symbolic goal description, followed by sampling-based methods to check whether the symbol-defined constraints can be satisfied. This limitation restricts the applicability of such methods to the tasks mentioned at the beginning. Moreover, it is worth noting that these tasks are, in fact, optimization problems rather than constraint satisfaction problems.

To address these issues, we introduce Logic-Skill Programming, an optimization-based approach designed to eliminate the need for explicit symbolic goal descriptions. It also provides a notion of global optimality over the full logic-geometric path. This work draws inspiration from Logic-Geometric Programming \cite{toussaint2015logic} and shares the same philosophy, with a key distinction that we concentrate on skill policy planning, rather than motion planning in a fully-known environment. 

Fig. \ref{fig:pipeline} is an overview of our proposed approach. Given the initial state $s_0$ and $\overline{\bm{x}}_0$, along with the evaluation function $\Psi$ for the final geometric configuration, LSP can generate the solutions to sequence multiple skills from the task-agnostic skill library. This problem is formulated as an extended first-order mathematical program, where first-order symbols are introduced to constrain the optimization problem. The objective is to optimize both the overall cumulative reward and the performance of the final geometric configuration. The overall cumulative reward is expressed as the sum of all value functions within the plan, forming a value function space. While Reinforcement Learning (RL) algorithms have excelled in tackling highly challenging problems, the approximations of the obtained value functions lack generalizability to the entire state space, which is critical in sequential skill planning for determining the optimal subgoal sequence with maximum cumulative reward. We therefore propose to use Tensor Train (TT) to approximate the value function space, which shows superior approximation capabilities compared to RL methods. The appropriate skill skeleton and optimal subgoal sequence are then obtained by alternating between symbolic search and skill value optimization over the value function space.

We summarize the contributions of this paper as follows:
\begin{itemize}
	\item We propose to formulate the sequential skill planning problem as an extended first-order mathematical program, eliminating the need for symbolic goal description and enabling to find the optimal solutions rather than only feasible ones.
	\item We propose Logic-Skill Programming to address sequential skill planning tasks by alternating between symbolic search and skill value optimization.
        \item We propose to use Tensor Train to approximate the value function space, aiding in finding the optimal subgoal sequence with maximum cumulative reward respecting system dynamics/kinematics.
\end{itemize}

\section{Related Work}


\subsection{Skill Learning}

Methods such as behavior cloning (BC) \cite{florence2022implicit} involve deep neural networks learning state-action distributions from offline datasets. This class of methods heavily relies on the quality of the provided datasets, lacking exploration in complex state spaces. An alternative is to consider RL. Thanks to the exploration-exploitation mechanism, RL algorithms can actively explore the state space and find a feasible path given any initial states. However, RL primarily explores the local state space determined by the exploration metric. Once the final target is reached, RL usually loses motivation to explore further to find more optimal solutions.

In this paper, we aim to find global optimality over the entire (logic and geometric) path. To achieve this, we need to know the value function accurately over the entire state space. Notably, both BC and RL methods fall short in providing this information. This assertion aligns with the motivation of Approximate Dynamic Programming (ADP), where the goal is to approximate the optimal value function for the entire state space. However, this objective is challenging due to the expensive storage and computation involved in solving the dynamic programming algorithms such as value iteration. We believe this challenge explains why existing literature primarily focuses on feasibility rather than pursuing optimality in sequential skill planning. Recently, Shetty et al. \cite{Shetty24ICLR} proposed the use of Tensor Train for ADP and performed well in several hybrid control benchmarks. We find it promising to extend this method to learning manipulation skills, leveraging its advantages for accurate value function approximation across the entire state space for optimal sequential skill planning.

\subsection{Hybrid Long-horizon Planning}

Robot manipulation is typically characterized by hybrid aspects, such as contact modes (sticking/sliding) and manipulation primitives (pushing/pivoting). This hybrid structure poses significant challenges for gradient-based optimization techniques. To address this issue, one class of methods uses MIP \cite{marcucci2019mixed}. MIP involves both discrete and continuous variables in optimization problems and relies on Branch-and-Bound techniques to efficiently search for optimal solutions by pruning undesirable ones. It performs well if the original problem can be reformulated as combinations of several convex optimization sub-problems. However, manipulation tasks are usually challenging to convexify. Another well-studied method is Mathematical Program with Complementary Constraints (MPCC), which eliminates integer variables by adding complementary constraints on the decision variables. Subsequently, augmented Lagrangian techniques are employed to solve constrained optimization problems. This class of methods has demonstrated excellent performance in legged robot locomotion \cite{posa2014direct} and planar manipulation tasks \cite{moura2022non}. However, the added constraints often render the problem more fragile and prone to getting stuck in poor local optima.

Another promising line of research is Task and Motion Planning (TAMP) \cite{garrett2021integrated}. It adopts the concept of MIP but replaces integers with symbols, enabling the utilization of powerful planning tools from the classical AI field. Existing methods in TAMP can be categorized into two branches: sampling-based and optimization-based methods. The sampling-based methods \cite{srivastava2014combined, garrett2020pddlstream} usually assume a symbolic goal description and rely on a high-level task planner to find a feasible action skeleton. These methods often introduce predicates to represent geometric properties at a symbolic level and aim to identify correct symbolic abstractions of geometries, allowing reasoning solely at the symbolic level. However, finding the right abstractions can be non-trivial. Conversely, optimization-based approaches \cite{toussaint2015logic, xue2023d} aim to solve problems at the geometric level, treating symbolic logic as constraints in mathematical programming. While these methods can be less efficient than sampling-based ones, they can handle arbitrary objectives specified only in terms of an evaluation function for the final geometric configuration. Moreover, they provide a notion of global optimality over the entire logic and geometric path. This philosophy has strongly influenced our work, inspiring the removal of symbolic goals and the pursuit of optimal solutions for sequential skill planning.

In general, TAMP performs well across various domains \cite{hartmann2022long, quintero2023optimal, yang2022sequence}. However, its reliance on full knowledge of the planning domain and dynamics model limits its applicability in realistic environments. For instance, modeling the contact or interaction between the robot and the environment is often impossible, leading to poor performance of offline planned trajectories in online setups. One potential solution to this issue could be the use of model predictive control for receding horizon planning \cite{toussaint2022sequence}. Nevertheless, this approach may result in somewhat short-sighted behavior, and the intrinsic combinatorial structure also makes it expensive in terms of online planning. The weaknesses of TAMP highlight the importance of leveraging learned skill policies to address uncertainties and disturbances in the real world, particularly those involving physical contact.

\subsection{Sequential Skill Planning}

Prior works that focus on sequential skill planning typically adopt the options framework \cite{sutton1999between}, where a high-level policy is trained to sequence low-level skills toward the final goals. For instance, Xu et al. \cite{xu2021deep} proposed learning a skill proposal network as the high-level policy and utilizes learned skill-centric affordances (value functions) to assess the feasibility of the proposed skill skeleton. Similarly, Shah et al. \cite{shah2021value} suggested using the value functions of low-level skills as the state space to represent the symbolic skill affordance. This representation is then utilized to train an upper-level RL policy toward long-horizon goals. Although such methods have demonstrated the ability to solve long-horizon tasks, the resulting policies are typically task-specific and exhibit poor generalization on unseen tasks. Moreover, we believe the capabilities of value function has not been fully explored by \cite{shah2021value, xu2021deep}. Instead of using the value function space symbolically, we regard it as an abstraction of cumulative reward, taking into account the system kinematics/dynamics in geometric level. Subsequently, we leverage it to identify the optimal subgoal sequence that results in the maximum cumulative reward of the full path.

To enhance generalization ability, several recent works rely on symbolic planning to sequence task-agnostic skills. In Agia et al. \cite{agia2023stap}, the product of Q-functions is maximized to ensure the joint success of all skills sequenced in a plan, with the skill skeleton given by a high-level task planner. Similarly, in Huang et al. \cite{huang2019continuous}, a symbolic planner is used to sequence imitation learning policies, by leveraging continuous relaxation to improve symbolic grounding success. In Wu et al. \cite{wu2021example}, a repertoire of visuomotor skills is learned through human-provided example images. The pre-conditions and effects of each action are represented by images, enabling the verification of the feasibility of the skill skeleton provided by a high-level symbolic planner. The key is the formulation of a constraint satisfaction problem where the effects of the parameterized skill primitive satisfy the preconditions of the next skill in the plan skeleton. The plan skeleton is either predefined \cite{mishra2023generative} or obtained through PDDL planners \cite{agia2023stap}, requiring an explicit symbolic goal description. None of these methods aims to optimize over a final configuration given only by an objective function, limiting their ability to solve the problem mentioned at the beginning.

Therefore, we are motivated to propose a method that eliminates the need for a symbolic goal description while ensuring the maximum cumulative reward of the obtained skill skeleton and motion trajectories. This work aligns with LGP \cite{toussaint2015logic}, except that we plan over skills rather than motions, as we believe skills are more applicable in realistic environments.

\section{Background}
\subsection{Tensor Train for Function Approximation}
\label{sec:back_TT}

\minew{A multivariate function $f(x_1,\ldots, x_d)$ over a rectangular domain $\mc{D}$ can be approximated by a tensor $\bm{\mc{F}}$, where each element in the tensor represents the value of the function given the discretized inputs. The value of function $f$ at any point in the domain can then be approximated by interpolating among the elements of the tensor $\bm{\mathcal{F}}$.}

\minew{Due to storage limitations, representing a high-dimensional tensor is challenging. Tensor Train (TT) decomposition \cite{oseledets2011tensor}  was proposed to solve this problem by representing the tensor using a set of third-order tensors called \textit{cores}. TT-Cross \citep{oseledets2010_ttcross1,savostyanov2011_ttcross2} is a widely used technique to approximate a function in TT format. For more introductions about these techniques, readers may refer to \cite{Shetty24ICLR} for a detailed review.}

\subsection{Logic-Geometric Program}
\label{sec:back_LGP}

Logic-geometric Programming was proposed by \cite{toussaint2015logic} to integrate first-order logic into a mathematical program for addressing combined Task and Motion Planning (TAMP) problems. The general formulation is as follows: Given a logic $\mathcal{L}$, a knowledge base $\mathcal{K} \in \mathcal{L}$, and an objective function $l(x)$ over geometric configurations $x \in \mathcal{X}$, the logic-geometric program can be expressed as:
\begin{equation}
	\min_{x, \kappa}  l(x) \quad 
	\textrm{s.t.}  \; \kappa \models \mathcal{K}, \; g(x, \kappa) \leq 0, \; h(x, \kappa) = 0, 
\end{equation}
where $\models$ represents logical implication, indicating that $\mathcal{K}$ is satisfied once the logical statement $\kappa$ is True. This defines the associated equality and inequality constraints: $g(x, \kappa) \leq 0$ and $h(x, \kappa) = 0$.

Our formulation can be viewed as a special case of logic-geometric program, where the objective function is the total cumulative reward and the evaluation of  the final configuration, and the knowledge base is the selected skill skeleton.

\section{Methodology}
\subsection{Problem Formulation}
\label{sec: problem_form}

Our objective is to address long-horizon manipulation tasks by sequentially executing a series of skills included in a skill library $\mathcal{P}=\{\pi_1, \pi_2, \dots, \pi_N \}$. To ensure that the learned skills can be sequenced in an arbitrary manner and generalized to any long-horizon tasks, the skills should be task-agnostic, as also described in Agia et al. \cite{agia2023stap}. This implies that the learned policy is trained independently, with its own skill-centric parameters. Each skill domain can be modeled by a Markov Decision Process (MDP)
\begin{equation}
\mathcal{M}_n = (\mathcal{X}_n, \mathcal{U}_n, \mathcal{T}_n, \mathcal{R}_n),
\end{equation}
where $\mathcal{X}_n$ is the state space, $\mathcal{U}_n$ is the action space, $\mathcal{T}_n (x'_n|x_n, u_n)$ is the transition model, $\mathcal{R}_n (x_n, u_n)$ is the reward function given current state and action. 

The full $K$-length long-horizon domain is the union of $\{\mathcal{M}_1, \mathcal{M}_2, \dots, \mathcal{M}_K \}$, where each time segment corresponds to a single skill policy. It is specified as
\begin{equation}
	\overline{\mathcal{M}}= (\mathcal{M}_{1:K}, \overline{\mathcal{X}}, \Phi_{1:K}, \Gamma_{1:K}),
\end{equation}
where $\overline{\mathcal{X}}$ is the full state space of the long-horizon domain, $\Phi_{1:K}: \mathcal{X}_k \rightarrow \overline{\mathcal{X}}$ is a function that maps the policy state space to the full state space, and $\Gamma_{1:K}: \overline{\mathcal{X}} \rightarrow \mathcal{X}_k$ is a function that extracts the policy state from the full state. Note that the state spaces of different skills usually have different dimensions and structures. Taking pushing and pivoting as examples, they are planar manipulation primitives defined in horizontal and vertical planes, respectively, affecting different elements of the object pose after execution. The long-horizon state space serves as a bridge, transmitting the local state changes from the previous skill to the subsequent one.

After acquiring these task-agnostic policies, our goal is to determine the optimal skill skeleton and subgoal sequence, given the evaluation function $\Psi$ of the final configuration. Each skill $\pi_k (\bm{u}|\bm{x})$ is trained to maximize the cumulative reward given the current state $\bm{x}$. Therefore, finding the optimal skill sequence aims to maximize the overall cumulative reward for the complete skill sequence $a_{1:K}$ and the evaluation of the final configuration $\Psi(\overline{\bm{x}}_{K_T})$:
\begin{equation}
\max_{\bm{x}, a_{1:K}}   \sum_{k=1}^{K} \mathbb{E}_{\pi_k} \left[ \sum_{t=0}^{\infty} \gamma^{t} R_{k_t} \right] + \Psi(\overline{\bm{x}}_{K_T}).
\label{eq:cum_rew}
\end{equation}

Here, $R_{k_t}$ denotes the reward of skill $a_k$ at time $t$. However, directly optimizing over the infinite full horizon is impossible. Considering that the value function is defined to approximate the cumulative reward in dynamic programming:
\begin{equation}
	V^{\pi_k}(\bm{x})=\mathbb{E}_{\pi_k} \left[ \sum_{t=0}^{\infty} \gamma^t R_{k_t} (\bm{x}_{k_t}, \pi_k(\bm{x}_{k_t})) \mid \bm{x}_{k_0} = \bm{x} \right],
\end{equation}
and the objective of each policy is to find an action from the action space that can maximize the cumulative reward from current state:
 \begin{align}
 	Q^{\pi_k}(\bm{x}, \bm{u}) &= \mathbb{E}_{\pi_k} \left[ \sum_{t=0}^{\infty} \gamma^t R_{k_t} (\bm{x}_{k_t}, \bm{u}_{k_t}) \mid \bm{x}_{k_0}\!=\! \bm{x}, \bm{u}_{k_0}\!=\!\bm{u} \right]\!, \nonumber\\
%
%
	\pi_k(\bm{x}) &= \arg\max_{\bm{u}} Q^{\pi_k}(\bm{x}, \bm{u}).
\end{align}

Sequential skill planning problem can be formulated as maximizing the sum of value functions, along with the evaluation function $\Psi$ of the final configuration, incorporating first-order logic as constraints:
\begin{equation}
\begin{aligned}
\max_{\overline{\bm{x}}, a_{1:K}, s_{1:K}}  &\sum_{k=1}^{K} V^{\pi_k}(\bm{x}_{k_0}) + \Psi(\overline{\bm{x}}_{K_T})  \label{eq:obj_func}  \\
\textrm{s.t.} \quad & s_k \in succ(s_{k-1}, a_k), 
\overline{\bm{x}}_{1_0} = \overline{\bm{x}}_0,\\
&h_{\text{path}}(\bm{x}_{k_t}, \pi_k(\bm{x}_{k_t}) | s_k) = 0, \\ & g_{\text{path}}(\bm{x}_{k_t}, \pi_k(\bm{x}_{k_t}) | s_k) \leq 0, \\
&h_{\text{switch}}(\overline{\bm{x}}_{k_T} | a_{k+1}, s_{k}) = 0     ,\\ & g_{\text{switch}}(\overline{\bm{x}}_{k_T} | a_{k+1}, s_{k}) \leq 0,\\
& \overline{\bm{x}}_{k_t} = \Phi_k (\bm{x}_{k_t}| a_k), \\
& \bm{x}_{k_0} = \Gamma_k (\overline{\bm{x}}_{k_0}, \overline{\bm{x}}_{k_T} | a_k),
\end{aligned}
\end{equation}
\minew{where $\overline{\bm{x}}_{k_0}$ and $\overline{\bm{x}}_{k_T}$ are the initial and final configuration of skill operator $a_k$ in the long-horizon domain $\mathcal{\overline{X}}$. The goal configuration of $a_k$ is identical to the initial configuration of $a_{k+1}$, namely $ \overline{\bm{x}}_{(k+1)_0} = \overline{\bm{x}}_{k_T}$. The initial state of $a_k$ within the skill domain $\mathcal{X}_k$, denoted as $\bm{x}_{k_0}$, is computed using $\overline{\bm{x}}_{k_0}$ and $\overline{\bm{x}}_{k_T}$. $V^{\pi_k}(\bm{x}_{k_0})$ therefore represents the cumulative reward from $\overline{\bm{x}}_{k_0}$ to $\overline{\bm{x}}_{k_T}$ in $\mathcal{X}_k$. $\Phi_k$ and $\Gamma_k$ are the mapping functions between long-horizon domain $\mathcal{\overline{X}}$ and skill domain $\mathcal{X}_k$. Thanks to $\Gamma_k$, given new initializations and targets, there is no need to retrain new value functions. The definition of $\Gamma_k$ is outlined in Sec. \ref{sec: ex_policy}.} $s_0$ and $\overline{\bm{x}}_0$ denote the initial symbolic state and geometric configuration, respectively. $h_{\text{path}}$ and $g_{\text{path}}$ indicate the constraints on the path $\bm{x}_{k_t}$ given current symbolic state $s_k$. Such constraints are addressed by the skill policies $\pi_k$ in each skill domain. $h_{\text{switch}}$ and $g_{\text{switch}}$ express the transition consistency of configuration $\overline{\bm{x}}_{k_T}$ with the following skill operator $a_{k+1}$. For example, to \texttt{pivot} an object against the wall, the object should be well-positioned in contact with the wall after the previous action. These constraints specify the degrees of freedom in the system configurations that can be actuated under the symbolic state $s_k$, thereby effectively reducing the number of decision variables in $\overline{\bm{x}}_{k_T}$ and streamlining the optimization process. For instance, an object marked as $\texttt{non-graspable}$ and $\texttt{onTable}$ can only be manipulated through $\texttt{push}$ or $\texttt{pull}$, with actuation in the dimensions of $\texttt{x}$, $\texttt{y}$ and \texttt{yall}. Conversely, if the object is marked as $\texttt{atWall}$, it can be actuated by $\texttt{pivot}$, with changes in $\texttt{roll}$ or $\texttt{pitch}$.

The value functions of all skills $a_{1:K}$ collectively constitute a value function space, offering insights into optimal control for sequentially executed skills, while considering system dynamics and kinematics. By optimizing over it, we can identify the subgoal sequence that maximizes the cumulative reward while satisfying system constraints. For instance, as depicted in Fig. \ref{fig:push}, the value function space can inform the planner which configuration along the table edge is more dynamically optimal, especially considering the under-actuated pushing dynamics.

We assume the initial geometric configuration $\overline{\bm{x}}_0$ is given, along with an initial symbolic state $s_0 \in \mathcal{L}$. The symbolic states can be transited through the skill operator $a_k$ as $s_k = succ(a_k, s_{k-1})$. The existing skill planning frameworks typically require an explicit symbolic goal target $s_T \models \mathcal{G}$, while our method eliminates this requirement by considering only the evaluation $\Psi$ of the final geometric configuration.

\subsection{Logic-Skill Programming}
\label{sec:LSP}

To solve Eq. \eqref{eq:obj_func}, we propose an approach that alternates between searching for symbolic skills and optimizing the value functions. The role of symbolic search is to define the optimization constraints, while skill value optimization checks the skeleton feasibility and evaluates the final configuration. Additionally, our method, instead of only seeking feasible solutions, aims to find the optimal solution that maximizes the overall cumulative reward while achieving the final target.


\subsubsection{Level 1: Symbolic Search}
\label{sec: sym_search}

We use Planning Domain Definition Language (PDDL) \cite{aeronautiques1998pddl} to describe the task domain and then utilize Monte Carlo Tree Search (MCTS) to search for the appropriate skill sequence $a_{1:K}$ from the skill library. The preconditions and effects of each skill form a rule-based representation of symbolic transitions $s_k = succ(s_{k-1}, a_k)$, serving as the forward model for symbolic search. In each iteration, MCTS relies on the Upper Confidence Bound (UCB1) \cite{Williams2016cognitive} to select the node:
\begin{equation}
    UCB1(s_k) = \frac{w_{s_k}}{v_{s_k}} + C_E \sqrt{\frac{2 \ln(v_{s^p_k})}{v_{s_k}}}, 
\end{equation}
where $v_{s_k}$ represents the number of visits on symbolic state $s_k$, and $w_{s_k}$ indicates the total accumulated reward obtained through state $s_k$. $v_{s^p_k}$ denotes the number of visits on the parent node of $s_k$. $C_E$ is a constant to balance exploitation and exploration.

If no child node is found in the current branch of the search tree, the branch will expand and simulate until it either reaches the target or surpasses the maximum length. Since our formulation does not have a symbolic goal, we rely on the skill value optimization process (Sec. \ref{sec: skill_opti}) to evaluate the performance of the final geometric configuration. A reward $\overline{r}$ will be backpropagated through the branch, depending on the simulation result. This iterative process refines the search tree, focusing on promising branches and ultimately converging towards an optimal decision. It is important to note that the sequence length ($K$ in Eq. \eqref{eq:obj_func}) is purely unknown before symbolic search, allowing diverse sequence lengths for the same target configuration. By applying a higher $C_E$ in the UCB1 formula, MCTS can return multiple solutions. For example, as shown in Fig. \ref{domain:ppm}, if we want to grasp the cube that is only graspable from the lateral side, multiple symbolic sequences can be found by symbolic search. One can be \texttt{push-pick}, pushing the cube to the edge and then grasping it from the table side, while another solution can be \texttt{push-pivot-pull-pick}. \minew{Given multiple solutions, we can either choose the most robust one before execution using domain knowledge, or switch between solutions during execution. This will be further studied in the future to fully exploit the advantage of multiple solutions.}
	
\subsubsection{Level 2: Skill Value Optimization}
\label{sec: skill_opti}

The symbolic skill skeleton can only express the feasibility in symbolic level. It has to be verified by Eq. \eqref{eq:obj_func} to evaluate the final configuration while satisfying the constraints. This verification is particularly crucial in our work, since the objective is solely defined by an evaluation function $\Psi$ of the final configuration. Additionally, since the sequenced skills are task-agnostic, finding the subgoal of each skill is crucial for the skill policy to reason about where to go given the current state. All of these considerations highlight the crucial necessity of solving Eq. \eqref{eq:obj_func}.

The first step involves obtaining the value functions that can accurately approximate the cumulative reward in each skill domain. To achieve this, our approach utilizes Tensor Train for value function approximation and skill policy learning, based on an algorithm called Generalized Policy Iteration using Tensor Train (TTPI) \cite{Shetty24ICLR}. It is an ADP method that aims to approximate the value function throughout the entire state space. TTPI has demonstrated great performance in hybrid control scenarios in \cite{Shetty24ICLR}, where the value function is used to compute the advantage function for policy retrieval. In this work, we apply TTPI to construct the value function space for skill sequencing.

After obtaining the optimal value functions for different manipulation skills, we can solve Eq. \eqref{eq:obj_func} conditioning on the skill skeleton generated from the symbolic search level. The resulting subgoals $\overline{\bm{x}}_{1_T:K_T}$ should satisfy both the path constraints and the switch constraints. The switch constraints $h_{\text{switch}}$ and $g_{\text{switch}}$ dictate that $\overline{\bm{x}}_{k_T}$ must lie within the intersection space of two adjacent skill domains, $\mathcal{X}_k$ and $\mathcal{X}_{k+1}$, ensuring configuration consistency. The path constraints $h_{\text{path}}$ and $g_{\text{path}}$ are addressed by the skill policies in each skill domain. For example, to verify whether the subgoal $\overline{\bm{x}}_{k_T}$ can be achieved while respecting path constraints, the skill policy $\pi_k$ should be executed in domain $\mathcal{X}_k$ given initial state $\bm{x}_{k_0}$. We assume that the difference between the model parameters used for skill policy learning and those in the real world is modest. Thanks to the receding-horizon mechanism during policy execution, the skill policy can achieve the target despite model uncertainties and disturbances, thereby naturally satisfying $h_{\text{path}}$ and $g_{\text{path}}$. This assumption is sensible, aligning with the objective of obtaining a powerful policy in the skill learning community and supported by the robust performance of learned task-specific policies in existing literature \cite{miki2022learning, Shetty24ICLR, chi2023diffusion}. The focus of our paper is then more about how to sequence multiple skills to accomplish a much more complicated task. Additionally, if this assumption does not hold, we can still use the learned skill policies and value functions to find the feasible solution set and select the optimal one that maximizes the objective function in Eq. \eqref{eq:obj_func}.

To design an appropriate optimization technique, it is essential to note that both discrete and continuous variables may be involved in this problem. For instance, in the task mentioned at the beginning, we have to rely on pivoting against the wall to change the $\texttt{roll}$ or $\texttt{pitch}$ angle of the box. This requires one face of the box to be parallel to the wall. In other words, the $\texttt{yall}$ angle of the box has to be in $[-\pi, -\pi/2, 0, \pi/2]$ after pushing. Moreover, \minew{the objective function can be in any form, either convex or non-convex,} depending on the value functions of selected skills. All of these factors pose significant challenges for optimization techniques. In this work, we employ the Cross-Entropy Method (CEM) \cite{rubinstein1999cross, de2005tutorial} with mixed distribution, namely CEM-MD, as the optimization technique. It can handle mixed-integer programming by using Gaussian distribution and categorical distribution for continuous and discrete variables, respectively. The distributions are iteratively updated towards the fraction of the population with higher objective scores until converging to the best solution. The pseudocode of CEM-MD is shown in Alg. \ref{alg:cem}. Note that the $C$ samples in each iteration are generated in batch (lines 5-10) for fast computation.

\begin{algorithm}
	\caption{CEM-MD: Cross-Entropy Method with Mixed Distribution}
	\begin{algorithmic}[1]
		\State \textbf{Input:} Initial mean $\bm{\mu}$, covariance matrix $\bm{\Sigma}$ for continuous variables, initial probability vector $\mathbf{p}$ for discrete variables, population size $C$, elite fraction $p$, maximum iterations $H$, number of categories $K_c$, skill sequence $a_{1:K}$, Initial configuration $\overline{{\bm{x}}}_0$, evaluation function $\Psi$
		\State \textbf{Output:} Subgoal sequence $\bm{x}^*$, cumulative reward: $J^*$
		\State Initialize $h \gets 0$
		\While{$h < H$}
		\For{$i = 1$ to $C$}
		\State Sample continuous variables: ${\bm{x}}_{\text{c}_i} \sim \mathcal{N}(\mu, \Sigma)$
		\State Sample discrete variables: ${\bm{x}}_{\text{d}_i} \sim \text{Categorical}(\mathbf{p})$
		\State Combine: ${\bm{x}}_i = ({\bm{x}}_{\text{c}_i}, {\bm{x}}_{\text{d}_i})$
            \State Evaluate the samples: $J_i \gets l({\bm{x}}_i, a_{1:K}, {\overline{\bm{x}}}_0, \Psi)$  
            
                \Comment{objective function in Eq. \eqref{eq:obj_func}}
		\EndFor
		
		\State Select the top $p$ solutions (elite set): $\{\hat{{\bm{x}}}_1, \hat{{\bm{x}}}_2, \ldots, \hat{{\bm{x}}}_{pC}\}$
		\State Update the mean and covariance for continuous variables:
		\[
		\mu \gets \frac{1}{pC} \sum_{i=1}^{pC} \hat{{\bm{x}}}_{c, i}
		\]
		\State Update the probability vector for discrete variables:
		\For{$k_c = 1$ to $K_c$}
		\[
		\mathbf{p}_{\text{elite}_{k_c}} = \frac{\text{count}(\hat{{\bm{x}}}_{d, k_c})}{pC}
		\]
		\EndFor
		\[
		\mathbf{p} \gets \mathbf{p}_{\text{elite}}
		\]
		\State $h \gets h + 1$
		\EndWhile 
		\State ${\bm{x}}_{\text{c}}^* \gets \mathcal{N}(\mu, \Sigma)$ and ${\bm{x}}_{\text{d}}^* \gets \text{Categorical}(\mathbf{p})$
		\State ${\bm{x}}^* = ({\bm{x}}_{\text{c}}^*, {\bm{x}}_{\text{d}}^*)$,
		 $J^* \gets l({\bm{x}}^*, a_{1:K}, {\overline{\bm{x}}}_0, \Psi)$
		\State \textbf{Output} ${\bm{x}}^*$ and $J^*$
	\end{algorithmic}
	\label{alg:cem}
\end{algorithm}

Overall, our proposed approach is to alternate between symbolic search and skill value optimization. The symbolic search is achieved by MCTS, with the obtained skill skeleton as the constraints for skill value optimization. The skill value optimization is achieved by CEM-MD, with its results to inform symbolic search. The values of visited nodes in the search tree will be updated, initiating a new iteration. Given an evaluation function $\Psi$ of the final configuration, this framework can return multiple solutions. The pseudocode of our Logic-Skill Programming method is shown in Alg. \ref{alg:lsp}.

\begin{algorithm}
	\caption{LSP: Logic-Skill Programming}
	\begin{algorithmic}[1]
		\State \textbf{Input:} Initial configuration $\overline{\bm{x}}_0$, evaluation function $\Psi$, initial symbolic state $s_0$, maximum iterations $\widetilde{H}$, maximum number of solutions $\widetilde{N}_s$
		\State \textbf{Output:} Skill skeleton $a_{1:K}$, Subgoal sequence $\overline{\bm{x}}_{1_T:K_T}$
		\State Initialize $h \gets 0$, $N_s \gets 0$, $\text{Value Set:} \; \mathcal{O}_v \gets \emptyset$, $\text{Solution Set:} \; \mathcal{O}_s \gets \emptyset$
		\While{$h < \widetilde{H}$}
		\State Symbolic search: $a_{1:K} = \text{MCTS}(s_0)$
		\State Skill value optimization: \\
		  \qquad \qquad                $\overline{\bm{x}}_{1_T:K_T}, J = \text{CEM-MD} (a_{1:K}, \overline{\bm{x}}_0, \Psi)$
		
		\If{$\text{solved}$}
                \State $N_s \gets N_s + 1$
			\State  $\mathcal{O}_v \gets \mathcal{O}_v \cup (J)$
			\State $\mathcal{O}_s \gets \mathcal{O}_s \cup (a_{1:K}, \overline{\bm{x}}_{1_T:K_T})$
		\EndIf
            \If{$N_s \geq \widetilde{N}_s$}
                \State $break$
            \EndIf
		\State $h \gets h + 1$
		\EndWhile 
		\State \textbf{Output} $\mathcal{O}_s$
	\end{algorithmic}
	\label{alg:lsp}
\end{algorithm}

\section{Experiments}
\label{sec:exp}
In this section, we compare our method with state-of-the-art baselines, in terms of policy learning, subgoal optimization, and sequential skill planning. 
\begin{figure*}[h]
	\centering
	\begin{subfigure}[b]{.3\linewidth}
		\centering
		\includegraphics[width=\linewidth]{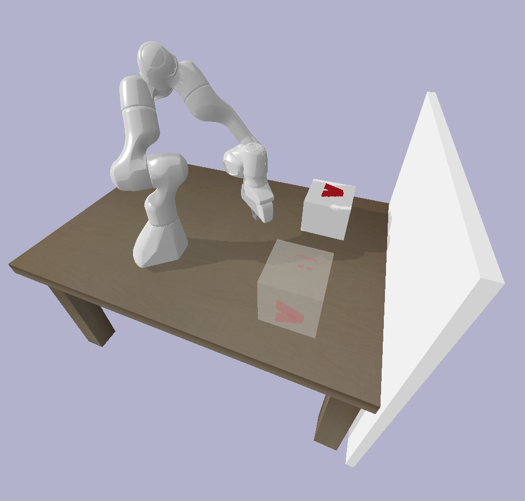}
		\caption{Non-Prehensile Manipulation domain}\label{domain:npm}
	\end{subfigure}\hfill
	\begin{subfigure}[b]{.3\linewidth}
		\centering
		\includegraphics[width=\linewidth]{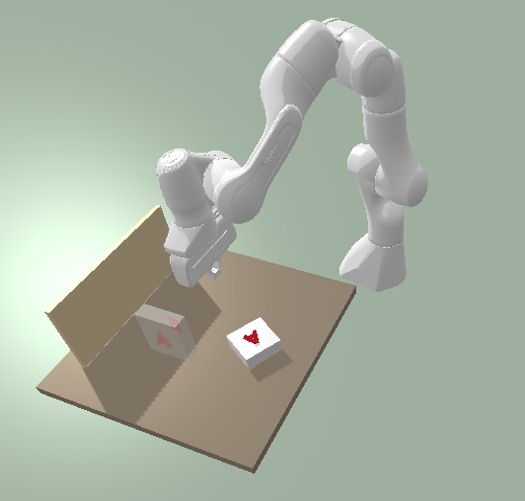}
		\caption{Partly-Prehensile Manipulation domain}\label{domain:ppm}
	\end{subfigure}\hfill
	\begin{subfigure}[b]{.3\linewidth}
		\centering
		\includegraphics[width=\linewidth]{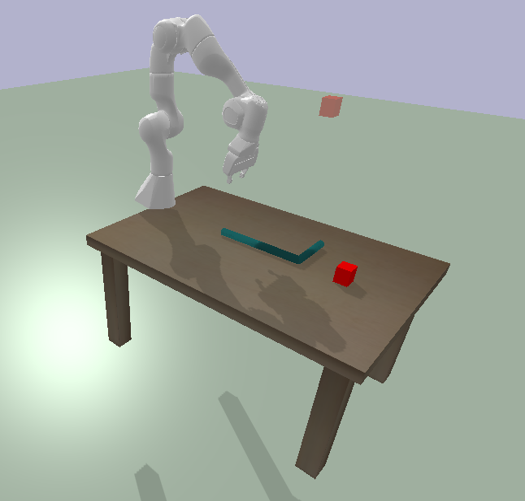}
		\caption{Prehensile Manipulation domain}\label{domain: pm}
	\end{subfigure}\hfill
	\caption{Three sequential manipulation domains, including both prehensile and non-prehensile manipulation primitives. The transparent object represents the final target configuration in each domain.}
	\label{fig:mani_domains}
\end{figure*}

\subsection{Evaluation on Skill Policy Learning}
\label{sec: ex_policy}

To evaluate the skill policies and their corresponding value functions, we initially construct a skill library using different skill learning methods. These methods include TTPI and two state-of-the-art RL methods: Soft Actor-Critic (SAC) \cite{haarnoja2018soft} and Proximal Policy Optimization (PPO) \cite{schulman2017proximal}.


		
	

\minew{The skill library encompasses five manipulation skills: pushing, pivoting, pulling, picking, and placing, each characterized by unique state and action spaces. Specifically, pushing, pivoting and pulling are planar manipulation primitives.}

\begin{enumerate}
\item \textbf{Push}: The state is characterized by $(\bm{p}_o, \theta_o, \bm{p}_r, f_c)$, while the action is denoted by $(\bm{v}_r, f_n)$. Here, $(\bm{p}_o, \theta_o) \in \text{SE(2)}$ denotes the object's pose in the world frame. $\bm{p}_r$ and $\bm{v}_r$ represent the position and velocity of the robot end-effector in the object frame. We assume that the object has a rectangular shape (common in industry), where $f_c \in {0,1,2,3}$ represents the current contact surface and $f_n$ denotes the next contact surface. Thus, the system encompasses a total of $6$ states and $3$ control variables, comprising both continuous and discrete variables.

\item \textbf{Pivot}: The pivoting domain features a goal-augmented state $(\beta, \tilde{\beta})$, where $\beta$ is the current rotation angle of the object in the gravity plane, and $\tilde{\beta}$ is the desired angle. The control input is $\dot{\beta}$, which denotes the angular velocity of the robot end-effector..

\item \textbf{Pull}: The state is defined as $(\bm{p}_o, \theta_o) \in SE(2)$, representing the object's position and orientation in a planar plane. The control input is $(\dot{\bm{p}}_o, \dot{\theta}_o)$, denoting the translational and angular velocities of the robot end-effector.

\item \textbf{Pick}: In this domain, the state is defined as $(x, y, z, \alpha, \beta, \theta) \in SE(3)$, representing the pose of the robot end-effector, while the control input is the Cartesian velocity of the end-effector, $(\dot{x}, \dot{y}, \dot{z}, \dot{\alpha}, \dot{\beta}, \dot{\theta}) \in SE(3)$. Without loss of generality, we assume that the object to be picked is located at $(0, 0, 0, 0, 0, 0)$.

\item \textbf{Place}: This domain shares the same state and action spaces as the \textbf{Pick} domain.

\end{enumerate}

We define the general reward for skill learning as: 
\begin{equation}
\begin{aligned}
    & r =  -1 \times (c_{p}+ \rho \times c_o + 0.01 \times c_a  + 0.1 \times c_f), \\
\end{aligned}
\end{equation}
with
\begin{equation}
\begin{aligned}
    & c_p = \lVert \bm{x_p} - \bm{x_p}^\text{des} \rVert/l_p ,\quad c_o = \lVert \bm{x_o} - \bm{x_o}^\text{des} \rVert/l_o , \\
    & c_a = \lVert \bm{u} \rVert,\quad c_f = 1 - \delta(f_c - f_n),
\end{aligned}
\end{equation}
where $\bm{x}_p$ and $\bm{x}_o$ represent the current position and orientation of the system in each skill domain, while $\bm{x}_p^\text{des}$ and $\bm{x}_o^\text{des}$ denote the target position and orientation. We set the state space to be within $[-0.5 m, 0.5 m]$ for position elements, and $[-\pi, \pi]$ for orientation elements. $l_p$ and $l_o$ are therefore set to $0.5$ and $\pi$ respectively to normalize the position error and orientation error. $\rho$ is a hyperparameter used to balance position and orientation errors. Due to the underactuated nature of pushing dynamics, we set $\rho=0.5$ in practice. For other skills, $\rho$ is set to $1$. $\bm{u}$ represents the control inputs of each skill domain. $c_f$ is a specialized term unique to the pushing domain, used to penalize face switching during pushing. Here, $\delta(f_c - f_n)$ returns 1 when $f_c=f_n$ (indicating no face switching), otherwise, it returns 0.

Notably, the skill policies learned for push, pull, pick and place are in a regulated manner, with the target state $\bm{x}^\text{des}$ set as $\bm{0}$. In terms of pivoting, the reward function does not include a positional term. In the orientation term, $\bm{x}_o$ is defined as $\beta$, and $\bm{x_o}^\text{des}$ is defined as $\tilde{\beta}$. Given new initial and target states in the long-horizon domain, along with the skill operator $a_k$, the skill-specific state can be computed as $\bm{x}_{k_0} = \Gamma_k (\overline{\bm{x}}_{k_0}, \overline{\bm{x}}_{k_T} | a_k)$ (as shown in Eq. \eqref{eq:obj_func}). For push, pull, pick and place, the domain mapping function $\Gamma_k$ is defined as:
\begin{equation}
\label{eq:domain_map1}
\Gamma_k(\overline{\bm{x}}_{k_0}, \overline{\bm{x}}_{k_T}) = \varphi_k(\overline{\bm{x}}_{k_0}) - \varphi_k(\overline{\bm{x}}_{k_T}),
\end{equation}
where $\varphi_k$ is a dimension reduction function that selects skill-specific dimensions from the long-horizon state. In the case of pivot, the skill is acquired through goal-augmented policy learning, achieved by augmenting the state as $(\beta, \tilde{\beta})$. This augmented state is the concatenation of the current rotation angle $\varphi_{\text{pivot}}(\overline{\bm{x}}_{k_0})$ with the desired angle $\varphi_{\text{pivot}}(\overline{\bm{x}}_{k_T})$. $\Gamma_{\text{pivot}}$ is therefore defined as:
\begin{equation}
\label{eq: domain_map2}
\Gamma_{\text{pivot}}(\overline{\bm{x}}_{k_0}, \overline{\bm{x}}_{k_T}) = \text{Concat}(\varphi_{\text{pivot}}(\overline{\bm{x}}_{k_0}), \varphi_{\text{pivot}}(\overline{\bm{x}}_{k_T})).
\end{equation}
The obtained $\bm{x}_{k_0}$ can then be directly fed into $V^{\pi_k}$ without the need to retrain new skill policies and value functions.

As for another domain mapping function $\Phi_k: \mathcal{X}_k \rightarrow \overline{\mathcal{X}}$ which maps skill-specific state to long-horizon state, it is defined as a function that updates the value of skill-specific dimensions in the long-horizon state based on the skill-specific state.


Notably, considering that SAC and PPO are not able to handle hybrid action space, we exclude $f_c$ and $f_n$ for a fair comparison of skill learning performance in this subsection. For both PPO and SAC, our primary implementation relies on Stable-Baselines3 \cite{stable-baselines3}, utilizing a Multilayer Perceptron (MLP) architecture with dimensions of $32 \times 32$ as the policy network. We set the discount factor to $0.99$ and the learning rate to $0.001$, while configuring the task horizon to $10^4$. In TTPI, we establish the accuracy threshold for TT-cross as $\epsilon=10^{-3}$ and set the maximum rank $r_{\max}$ to $10^2$.

The obtained policies are then evaluated by comparing the success rates across 1000 initial states in skill domains. \minew{We define a successful task as achieving a position error of less than 0.03cm and an orientation error of less than $15^\circ$.} Table \ref{tab:policy_eva} reveals that all the policies can nearly reach the final targets, demonstrating the proficiency of the learned policies in accomplishing individual manipulation tasks. However, rather than focusing solely on having skills for each specific task, we are also concerned with finding the optimal trajectory with the maximum cumulative reward for the sequenced skills. To achieve this, an accurate value function is required to inform which state in the intersection space between two adjacent skills leads to the maximum cumulative reward. Therefore, we examine whether the value functions can offer the same guidance as the cumulative reward given two states, shown as $\textit{value prediction}$ in Table \ref{tab:policy_eva}. The metric is to compare whether $V^{\pi_k}(\bm{x}_1) - V^{\pi_k}(\bm{x}_2)$ has the same direction as $\mathcal{R}^{\pi_k}_c(\bm{x}_1) - \mathcal{R}^{\pi_k}_c(\bm{x}_2)$, where $\mathcal{R}^{\pi_k}_{c}(\bm{x})$ is the cumulative reward, defined as
\begin{equation}
	 \mathcal{R}^{\pi_k}_c(\bm{x})=\sum_{t=0}^{\infty} \gamma^t R_{k_t} (\bm{x}_{k_t}, \pi_k(\bm{x}_{k_t})), \; \text{s.t.} \quad \bm{x}_{k_0} = \bm{x}.
\end{equation}

We randomly selected 1000 state pairs in the state domain. The value function of TTPI demonstrates superior prediction performance compared to SAC and PPO. This indicates that value functions in TT format offer better accuracy in approximating the cumulative reward, which can inform which state in the state domain is more dynamically optimal given current policy. This observation aligns with our motivation, emphasizing that typical RL methods approximate the value function primarily within a local space and the policy retrieval is often sub-optimal due to gradient-based optimization. In contrast, TTPI, as an ADP method, aims to approximate the value function across the entire state space. This feature is advantageous in our sequential skill planning framework for identifying the subgoal for each skill, which can be anywhere within the skill-specific state space.


\renewcommand{\arraystretch}{1.}
\begin{table*}[htbp]
	\centering    

	\caption{Comparative Analysis of Skill Policy Performance and Value Function Precision}
	\begin{footnotesize}
			\begin{tabular}{l |c c c| c c c| c c c|}
				\toprule
				& \multicolumn{3}{c|}{TTPI} & \multicolumn{3}{c|}{SAC} & \multicolumn{3}{c|}{PPO}\\
				& success rate & value prediction & \minew{time (min)}  & success rate & value prediction &\minew{time (min)}   & success rate & value prediction &\minew{time (min)} \\
				\cline{1-10}
				{pushing} &1.0 &0.85  &\minew{5.6} & 0.83  &0.63 &\minew{36.3} &0.95 &0.61 &\minew{44.8}\\
				{pivoting} &1.0  &0.94 &\minew{0.9} &1.0   &0.51 &\minew{16.0} &1.0 &0.68 &\minew{8.7} \\
				{pulling} &1.0   &0.97 &\minew{1.1}  &1.0    &0.84 &\minew{16.5} &1.0    &0.72 &\minew{10.7}  \\
				{pick/place} &1.0  &0.93 &\minew{1.67}  &0.98   &0.64 &\minew{33.6} &1.0   &0.66 &\minew{24.6}\\
	
				\bottomrule
			\end{tabular}
		
	\end{footnotesize}
	\label{tab:policy_eva}
	
\end{table*}

\subsection{Evaluation on Skill Value Optimization}
\label{sec:ex_opti}

\renewcommand{\arraystretch}{1.}
\begin{table*}[htbp]
		\centering    
	\caption{Comparison of Computation Error and Time for Skill Value Optimization}
		\begin{tabular}{l |c c c | c c| c c| c c|}
			\toprule
			& \multicolumn{3}{c|}{TTGO} & \multicolumn{2}{c|}{Shooting} & \multicolumn{2}{c|}{CEM-MD}\\ 
			& error & approximation time (s) & inference time (s) & error & time (s) & error & time (s)\\ 
			\cline{1-8}
			{NPM} &$0.03 \pm 0.00$ &$0.48 \pm 0.21$ &$0.003 \pm 0.00$  &$0.35 \pm 0.01$ &$0.01 \pm 0.00$   &$0.02 \pm 0.01$    &$0.06 \pm 0.01$\\ 
			{PPM} &$0.12 \pm 0.01$  &$3.90 \pm 0.22$  &$0.004 \pm 0.00$   &$0.59 \pm 0.01$ &$0.02 \pm 0.00$  &$0.05 \pm 0.01$    &$0.09 \pm 0.01$ \\
			{ PM} &$0.12 \pm 0.01$ &$20.28 \pm 5.23$ &$0.01 \pm 0.01$  &$0.75 \pm 0.27$ &$0.01 \pm 0.00$  &$0.03 \pm 0.01$    &$0.10 \pm 0.01$ \\  

			\bottomrule
		\end{tabular}
		
	\label{tab:opti_eva}
	
\end{table*}

\begin{figure*}[h]
	\centering
	\begin{subfigure}[b]{.3\linewidth}
	\centering
		\includegraphics[width=\linewidth]{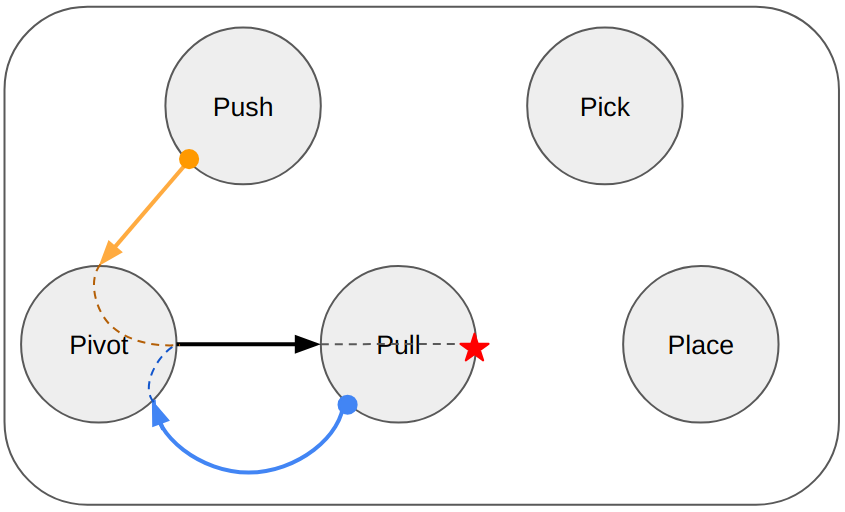}
		\caption{Non-Prehensile Manipulation}\label{lsp:npm}
	\end{subfigure}\hfill
	\begin{subfigure}[b]{.3\linewidth}
	\centering
	\includegraphics[width=\linewidth]{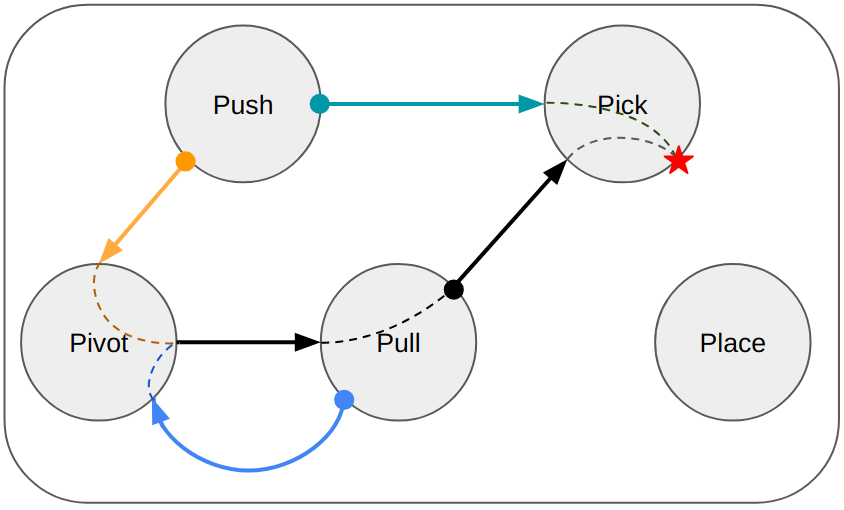}
	\caption{Partly-Prehensile Manipulation}\label{lsp:ppm}
	\end{subfigure}\hfill
	\begin{subfigure}[b]{.3\linewidth}
		\centering
		\includegraphics[width=\linewidth]{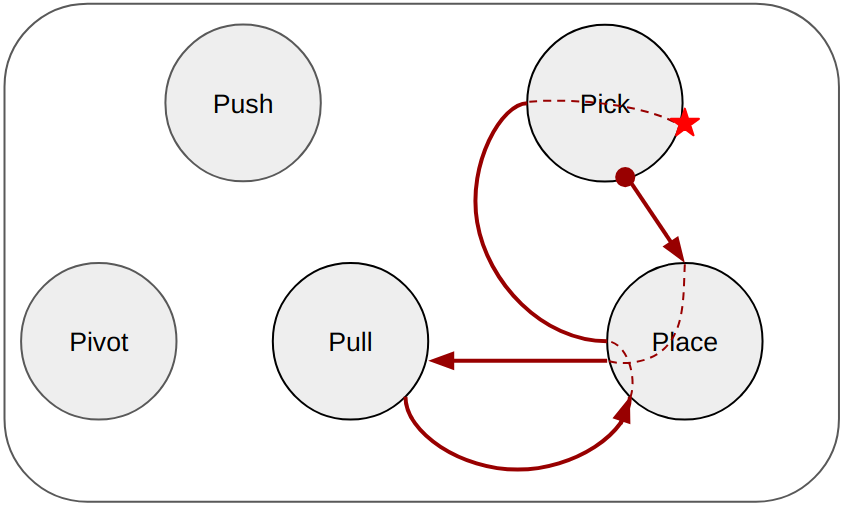}
		\caption{Prehensile Manipulation}\label{lsp:pm}
	\end{subfigure}\hfill
	\caption{The action skeletons obtained by LSP for three domains. The dot point denotes the start of the skill sequence. Each color represents one solution, with black lines indicating the common shared tunnel. The red star illustrates the end of the skill skeleton.}
	\label{fig:mcts}
\end{figure*}

\renewcommand{\arraystretch}{1.}
\begin{table*}[htbp]
	\centering    
	\caption{Comparison of Computation Time and Solution Quality for Sequential Skill Planning}
		\begin{tabular}{l |c c | c c| c c|c c|}
			\toprule
			& \multicolumn{2}{c|}{Computation Time (s) [w/ sym. goal]} & \multicolumn{2}{c|}{Computation Time (s) [w/o sym. goal]} & \multicolumn{2}{c|}{Cumulative Reward} &  {Sequence  Length} \\
			& LSP & STAP  & LSP & STAP & LSP & STAP \\
			\cline{1-8}
			{NPM} &$0.14 \pm 0.23$  &$0.06 \pm 0.03$  &$0.27 \pm 0.15$   &NA &$3.0 \pm 0 $ &$2.4 \pm 1.57 $ &$3.0 \pm 0$\\
			{PPM} &$0.17 \pm 0.1$   &$0.08 \pm 0.02$ &$0.22 \pm 0.13$      &NA &$2.9 \pm 0.7$ &$1.88 \pm 1.01$ &$2.9 \pm 0.7$ \\
			{ PM} &$0.25 \pm 0.15$  &$0.21 \pm 0.02$ &$0.41 \pm 0.02$     &NA &$5.0 \pm 0$ &$3.28 \pm 0.62$  &$5.0 \pm 0$ \\
			
			\bottomrule
		\end{tabular}
		
	\label{tab:lsp_eva}
	
\end{table*}

Table \ref{tab:policy_eva} illustrates that the acquired skills are successful in accomplishing each specific manipulation task. However, to achieve a long-horizon task where only the evaluation function $\Psi$ of final configuration is given, finding the subgoal for each skill is important. This necessitates a good optimization technique capable of finding the optimal solution based on the given objective function. 

Three different long-horizon domains are used to validate our method, involving both non-prehensile and prehensile manipulation primitives, as shown in Fig. \ref{fig:mani_domains}. The simulation environments are built using PyBullet \cite{coumans2019} and the AIRobot library \cite{airobot2019}.

\textbf{a) Non-Prehensile Manipulation (NPM)}: As depicted in Fig. \ref{domain:npm}, this domain involves objects that cannot be grasped. The objective is to manipulate the box within the 3D world by leveraging multiple non-prehensile planar manipulation primitives and establishing contacts with the surroundings to achieve the final 6D pose. This task can also be seen as a special case of in-hand manipulation, where the robot and the wall act as active and passive "fingers", respectively.

\textbf{b) Partly-Prehensile Manipulation (PPM)}: As shown in Fig. \ref{domain:ppm}, this domain involves objects that can only be grasped in specific directions. The goal in this domain is to manipulate the 6D pose of the cube, including the $z$ direction. Therefore, the robot must strategically decide how to pick up the object in the end. This domain requires the robot to reason about physical contact and object geometry to unify both prehensile and non-prehensile primitives.

\textbf{c) Prehensile Manipulation (PM)}: As illustrated in Fig. \ref{domain: pm}, this domain involves objects that can be directly grasped. The goal is to alter the 6D pose of a block placed on the table, beyond the robot's reachability. To achieve this, the robot must pick up the block in the end. This requires the robot to figure out using a hook to extend the kinematic chain and pull the block back into the reachability region, and then grasp it. However, the way in which the robot picks up the hook will affect whether the block can be reached, and where to pull will also affect the entire trajectory and the total energy cost. In this task, we want to show that our method can find an optimal trajectory which has the minimum energy cost, while successfully finishing the task.

\minew{We therefore define the evaluation function $\Psi$ of the final configuration $\overline{\bm{x}}_{K_T}$ as follows:
\begin{equation}
    \Psi(\overline{\bm{x}}_{K_T}) = \lambda \lVert \overline{\bm{x}}_{K_T} - \overline{\bm{x}}_{T} \rVert, 
\end{equation}
where $\overline{\bm{x}}_{T}$ is the target configuration in each domain, and $\lambda=10^2$. The initial configuration $\overline{\bm{x}}_0$ and target configuration $\overline{\bm{x}}_T$ are randomly sampled from the long-horizon domain $\overline{\mathcal{X}}$, which is the configuration space of the entire environment. In NPM and PPM, $\overline{\mathcal{X}} = \mathcal{S}_R \times \mathcal{S}_O$, while in PM, $\overline{\mathcal{X}} = \mathcal{S}_R \times \mathcal{S}_O \times \mathcal{S}_T$. Here, $\mathcal{S}_O = SE(3)$ and $\mathcal{S}_R = SE(3)$ denote the pose of the object and robot end-effector, respectively, while $\mathcal{S}_T = SE(2)$ represents the pose of a tool positioned on the table.}

It is worth noting that both PPM and NPM tasks involve optimizing both discrete and continuous variables, akin to mixed-integer programming. This complexity presents significant challenges in optimization, especially considering that the objective functions are arbitrary without any structure, depending on the value functions of skills. To this end, we compare CEM-MD with some other techniques capable of handling mixed-integer variables, including TTGO \cite{Shetty23} and the random shooting method. TTGO is an optimization technique specifically designed for functions in tensor train format, enabling it to find the optimal solution extremely fast once the TT model is provided. Random shooting is a naive technique that randomly samples variables from the domain and returns the one with the highest objective score. We apply these three optimization techniques across the three domains, conducting 10 random initializations for each domain. \minew{The maximum number of iterations for CEM-MD is set to $H=300$, with an early stopping criterion of $10^{-3}$ to terminate the while loop if the objective value no longer decreases. The population size is set to $C = 1000$, and the elite fraction is set to $p = 0.3$. The number of categories $K_c$ is set to $4$. The initial parameters $\bm{\mu}$, $\bm{\Sigma}$, and $\mathbf{p}$ are computed using samples collected from a uniform distribution in the first iteration.} The results are shown in Table \ref{tab:opti_eva}, with the computation error defined as the L2 norm between the final state $\overline{\bm{x}}_{K_T}$ and the target configuration $\overline{\bm{x}}_T$.

Given an arbitrary function, TTGO needs to approximate it first using TT-cross and then optimize over the approximated TT model, corresponding to the approximation stage and inference stage, separately. Table \ref{tab:opti_eva} shows that the approximation stage usually takes a longer time, but once the TT model has been obtained, the inference stage will be very fast. This suits problems with static objective functions quite well \cite{Shetty23}. However, in this work, if the high-level skill skeleton $a_{1:K}$ changes, the resulting objective function will change as well. This requires a new TT approximation, leading to expensive computation. Moreover, TTGO requires discretization on each dimension, resulting in a sub-optimal solution if no fine-tuning stage is used afterward. Meanwhile, the shooting method is much faster but can obtain poor solutions because of the random distribution. CEM-MD takes a bit longer time compared to random shooting but obtains good solutions by updating the mixed continuous and discrete distribution. The total computation time for these three domains is less than 0.1s, which is still fast enough in terms of long-horizon planning.
 
\subsection{Evaluation on Overall Performance of LSP}
\label{sec: ex_lsp}

\renewcommand{\arraystretch}{1.}
\begin{table*}[htbp]

\centering    
{
	\caption{\minew{The Specification of Skill Operators, Preconditions and Effects. In the effects column, only the states that are changed after skill execution are listed. The remaining symbolic states in the preconditions will be directly inherited by the effects.}}
		\begin{tabular}{c c c}
			\toprule
			Skill Operator & Preconditions & Effects \\ 
			\cline{1-3}
			\rowcolor{yellow!30} {push\_wall} &\text{$(\lnot$ (AtEdge $o$)) $\land$ $(\lnot$ (AtWall $o$)) $\land$ $(\lnot$ (AfterFlip $o$)) $\land$ (onTable $o$)}   &\text{(AtWall $o$)} \\
			\rowcolor{yellow!30} {pivot} &\text{(AtWall $o$)}  &\text{(AtWall $o$)  $\land$  (AfterFlip $o$)} \\ 
                \rowcolor{yellow!30} {pull\_wall} &\text{$(\lnot$ (AtEdge $o$)) $\land$ $(\lnot$ (AtWall $o$)) $\land$ $(\lnot$ (AfterFlip $o$))  $\land$ (onTable $o$)}   &\text{(AtWall $o$)} \\
			\rowcolor{yellow!30} {pull\_center} &\text{(AtWall $o$)  $\land$  (AfterFlip $o$)  $\land$ (onTable $o$)} &\text{($\lnot$ (AtWall $o$))}\\
                \rowcolor{gray!30} {pull\_edge} &\text{($\lnot$ (AtEdge $o$)) $\land$ $(\lnot$ (AtWall $o$)) $\land$ ($\lnot$ (AfterFlip $o$))  $\land$ (onTable $o$)}   &\text{(AtEdge $o$)} \\
                \rowcolor{gray!30} {pick\_edge} &\text{(AtEdge $o$) $\land$ (PartGraspable $o$) $\land$  (HandEmpty $r$) }   &\text{(InHand $o$) $\land$ ($\lnot$ (HandEmpty $r$))} \\
                \rowcolor{gray!30}{pick\_center} &\text{($\lnot$(AtWall $o$)) $\land$ (AfterFlip $o$) $\land$ (PartGraspable $o$) $\land$  (HandEmpty $r$) }   &\text{(InHand $o$) $\land$ ($\lnot$ (HandEmpty $o$))} \\
                \rowcolor{gray!30} {push\_edge} &\text{($\lnot$ (AtEdge $o$)) $\land$ $(\lnot$ (AtWall $o$)) $\land$ ($\lnot$ (AfterFlip $o$))  $\land$ (onTable $o$)}   &\text{(AtEdge $o$)} \\
                \rowcolor{green!30} {pick\_tool} &\text{($\lnot$ (ReadyPull $t$ $o$)) $\land$ (Graspable $o$) $\land$ (Reachable $t$) $\land$  (HandEmpty $r$)}   &\text{($\lnot$ (HandEmpty $r$))} \\
                \rowcolor{green!30}  {pull\_tool} &\text{(ReadyPull $t$ $o$) $\land$ ($\lnot$ (Reachable $o$)) $\land$ ($\lnot$ (HandEmpty $r$))  $\land$ (onTable $o$)}  &\text{(Reachable $o$)} \\
                \rowcolor{green!30}  {place\_toolmove} &\text{($\lnot$ (HandEmpty $r$)) $\land$ ($\lnot$ (Reachable $o$))}   &\text{(ReadyPull $t$ $o$)} \\
                \rowcolor{green!30}  {place\_tool} &\text{($\lnot$ (HandEmpty $r$))}   &\text{(HandEmpty $r$)} \\
                \rowcolor{green!30}  {pick\_object} &\text{(Reachable $o$) $\land$ (Graspable $o$) $\land$ (HandEmpty $r$)}   &\text{(InHand $o$)} \\

			\bottomrule
   \label{tab:skill_ope}
		\end{tabular}}
		
\end{table*}

\renewcommand{\arraystretch}{1.}
\begin{table*}[htbp]
		\centering    
	\caption{\minew{Initial Symbolic State and Symbolic Goal of Each Domain. Note that LSP does not need symbolic goals. Such goals are needed by the sampling-based sequential skill planning methods for comparison.}}
		\begin{tabular}{c c c}
			\toprule
			Domain & Initial symbolic state & \color{gray}{Symbolic goal} \\ 
			\cline{1-3}
			{NPM} &\text{\tabincell{c} {$(\lnot$ (AtEdge $o$)) $\land$ $(\lnot$ (AtWall $o$))  $\land$ $(\lnot$ (AfterFlip $o$))}  $\land$ (onTable $o$)} & \color{gray} \text{(AfterFlip $o$) $\land$ ($\lnot$ (AtWall $o$))} \\
			{PPM} &\text{\tabincell{c} {$(\lnot$ (AtEdge $o$)) $\land$ $(\lnot$ (AtWall $o$))  $\land$ $(\lnot$ (AfterFlip $o$))  $\land$  (HandEmpty $r$) $\land$ (PartGraspable $o$)  $\land$ (onTable $o$) }} &\color{gray} \text{(Inhand $o$)} \\ 
			{PM}  &\text{\tabincell{c} {($\lnot$ (Reachable $o$)) $\land$ (Graspable $o$) $\land$ (Reachable $t$)  $\land$  (HandEmpty $r$)  $\land$ (onTable $o$)}} &\color{gray} \text{(Inhand $o$)}\\
			\bottomrule
   	\label{tab:init_sym}
		\end{tabular}

	
\end{table*}

We then run the complete LSP framework for these three domains to assess the overall performance. \minew{In MCTS, the exploration parameter $C_E$ is set to 3, allowing for efficient tree search while exploring different skeleton solutions. The feedback reward $\overline{r}$ is defined as a binary variable $0/1$, determined by whether the geometric target is achieved. The maximum iteration is set to $\widetilde{H} = 100$, with an early stopping criterion that terminates the while loop upon finding enough $\widetilde{N}_s$ solutions. We set $\widetilde{N}_s=5$ to explore multiple solutions. Note that the $\widetilde{N}_s$ solutions can be identical, depending on MCTS and the total number of feasible solutions. We derive 13 skill operators from the five skill policies existing in the skill library. The logic defining the preconditions and effects of these operators is presented in Table \ref{tab:skill_ope}. Each operator's name, preceding the underscore, denotes the skill policy it utilizes. Elements highlighted in yellow are common across both NPM and PPM domains, those in gray are specific to the PPM domain, and those in green are specific to the PM domain. Throughout the table, we use the following symbols: $o$ for an object, $t$ for a tool, and $r$ for a robot arm. Table \ref{tab:init_sym} displays the initial symbolic state $s_0$ used in each domain, along with the symbolic goals used for comparison.}


\begin{figure}[t]
	\centering
	\begin{subfigure}[b]{0.23\textwidth}
		\includegraphics[width=\textwidth]{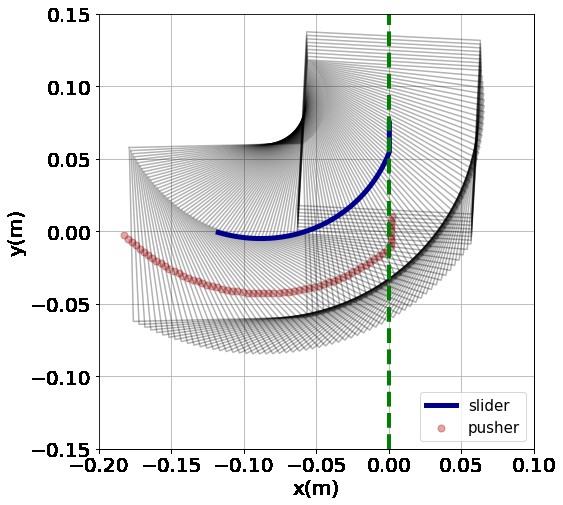}
		\caption{LSP}
		\label{sub:optimal}
	\end{subfigure}
	\begin{subfigure}[b]{0.23\textwidth}
		\includegraphics[width=\textwidth]{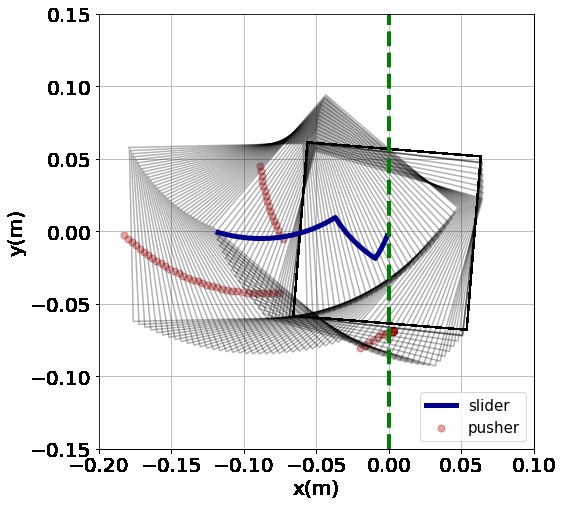}
		\caption{STAP}
		\label{sub:feasible}
	\end{subfigure}
	\caption{The pushing subtask obtained for the PPM domain. Both \ref{sub:optimal} and \ref{sub:feasible} have the same initialization configuration. The objective is to employ the robot end effector (pusher) to move the slider to the green line ($x=0$), representing the edge of the table. LSP provides a solution with the highest value in the space, requiring less control effort, while STAP outputs one that involves multiple face switches.}
	\label{fig:push}
\end{figure}

\minew{Given the same initial state $s_0$ and $\overline{\bm{x}}_0$ in each domain, we randomly sample 10 different target configurations $\overline{\bm{x}}_T$ that require multi-step manipulation.  Fig. \ref{fig:mcts} shows that LSP can actively find multiple solutions.} For the Non-Prehensile domain, two solutions found by LSP are \texttt{push-pivot-pull} and \texttt{pull-pivot-pull}. For the Partly-Prehensile domain, four different skeletons are found: \texttt{push-pick}, \texttt{pull-pick}, \texttt{push-pivot-pull-pick} and \texttt{pull-pivot-pull-pick}. For the Prehensile domain, the solved skill skeleton is \texttt{pick-place-pull-place-pick}.

We then compare our method with another state-of-the-art sampling-based sequential skill planning method called STAP \cite{agia2023stap}. STAP focuses on constraints satisfaction. It requires an explicit symbolic goal for high-level task planning, followed by feasible solutions sampling. \minew{To ensure a fair comparison, we use MCTS with an explicit symbolic goal as the task planner in STAP and then employ CEM-MD for feasibility checking given the skill skeleton.} Table \ref{tab:lsp_eva} illustrates the time required to find one solution and the solution quality between LSP and STAP. We can observe that LSP does not require a symbolic target goal, whereas STAP relies on it. STAP can find the solution faster than LSP. The reason is that STAP focuses only on finding the feasible solution, while LSP aims to provide (global) optimality over the full logic-geometric path. This aligns with the comparison of cumulative rewards. Note that the cumulative reward of each skill in the sequence is normalized by the highest value in the corresponding value function. We can observe that the trajectory found by LSP leads to a higher cumulative reward compared with STAP, indicating better optimality. Moreover, in the PPM domain, the skill sequence varies with different lengths, showing that multiple solutions are found, as depicted in Fig. \ref{lsp:ppm}.

Fig. \ref{fig:push} illustrates a toy example of the solutions found by LSP and STAP for the PPM domain, where the robot needs to push the block (slider) to the edge of the table (depicted as the green line) for grasping from the lateral side. LSP determines the subgoal with the highest cumulative reward by utilizing the value function, which represents the most optimal state considering the system dynamics. In contrast, STAP outputs a feasible state, which can be any configuration on the edge theoretically. Here, we pick up the one closest to the initialization in Euclidean space. However, to reach this target, the pusher exerts more effort, involving two face switches (visible in the discontinuous pusher trajectory in Fig. \ref{fig:push}). This also underscores the utility of the value function space for finding the optimal path considering system dynamics, compared with Euclidean state space.

\subsection{Real-robot Experiments}
\label{sec: ex_real}

\begin{figure*}[t] 
	\centering
	\subfloat[Initialization]{{\includegraphics[width=0.5\columnwidth]{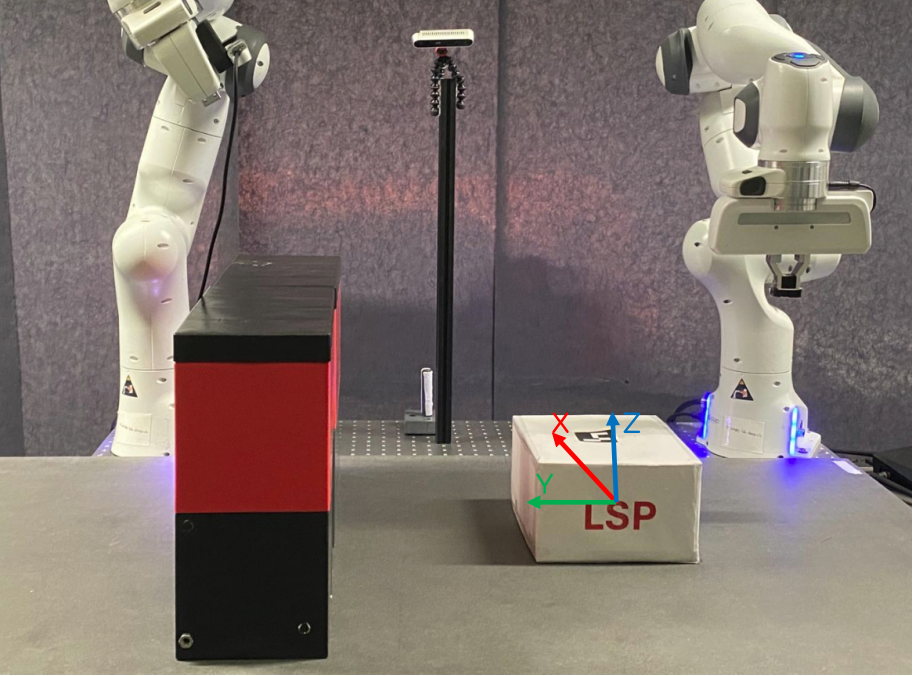}}\label{fig:robot_init}}
	\subfloat[Pushing]{{\includegraphics[width=0.5\columnwidth]{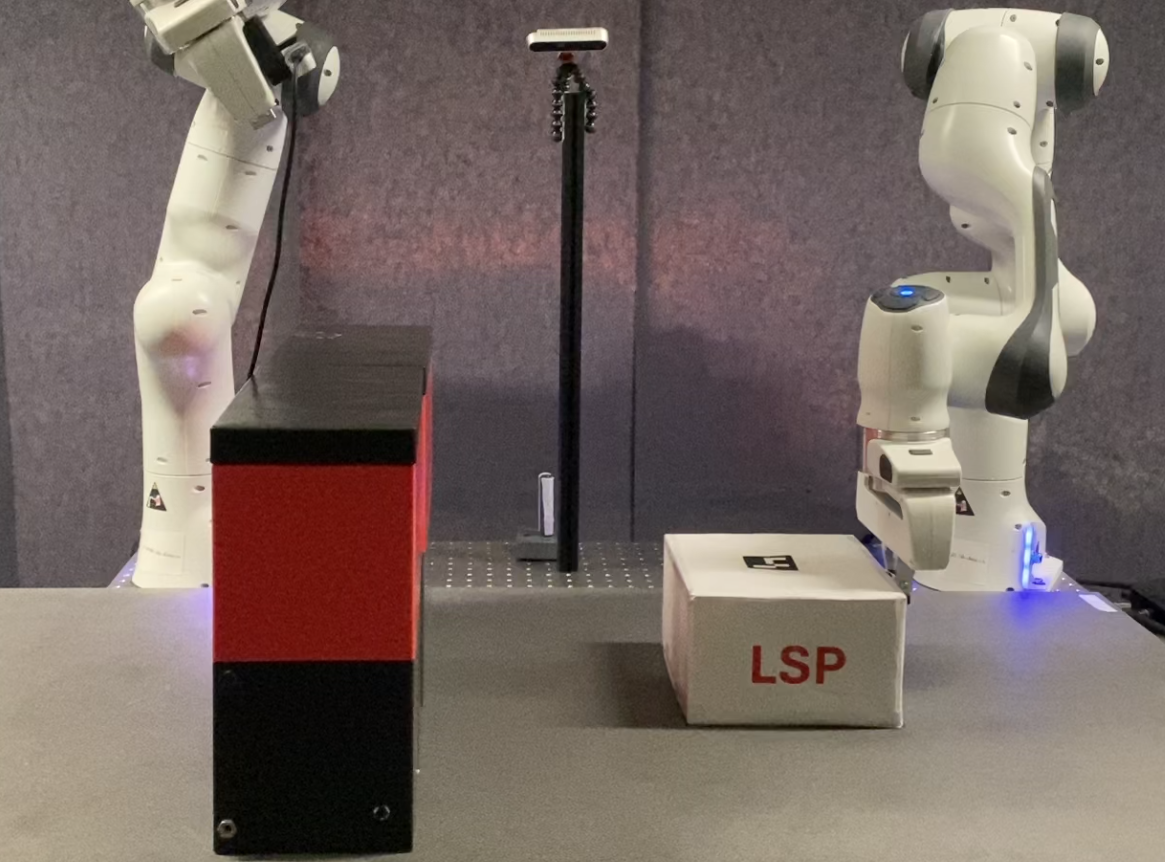}}} 
	\subfloat[Disturbance]{{\includegraphics[width=0.52\columnwidth]{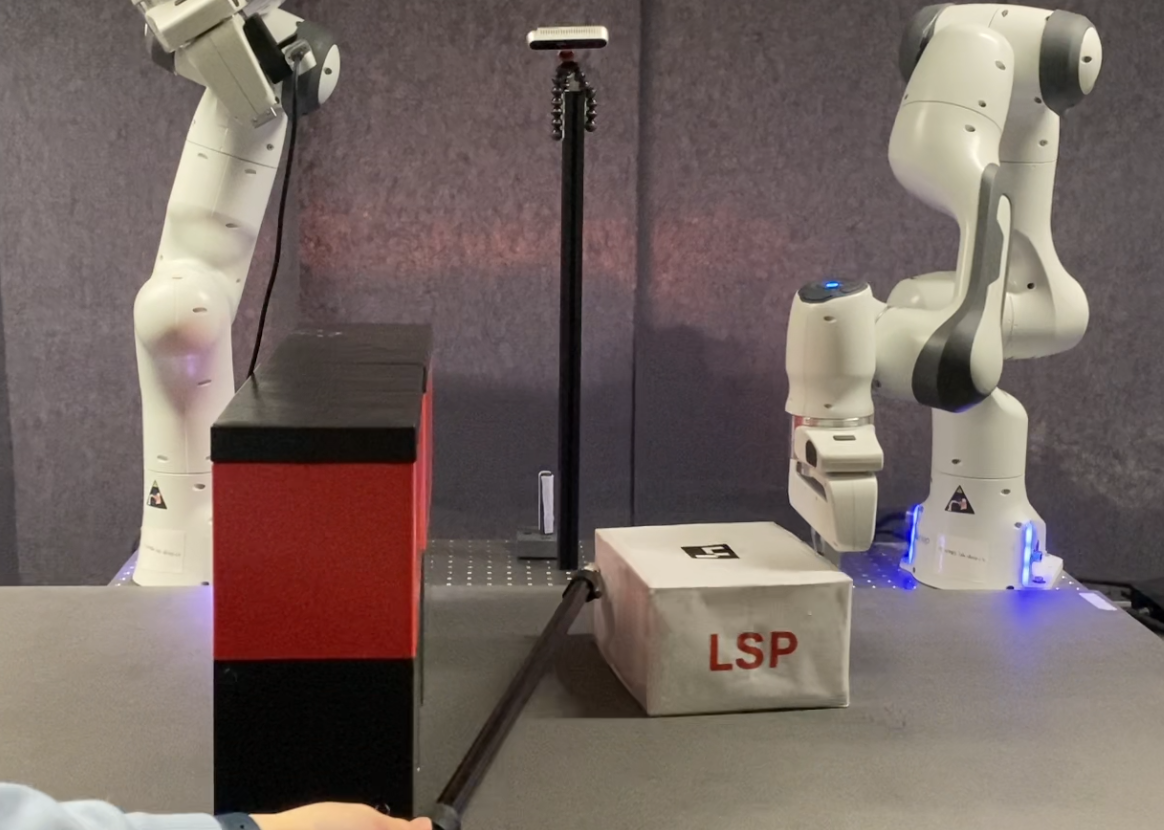}}} 
	\subfloat[Push-Pivot Switch]{{\includegraphics[width=0.5\columnwidth]{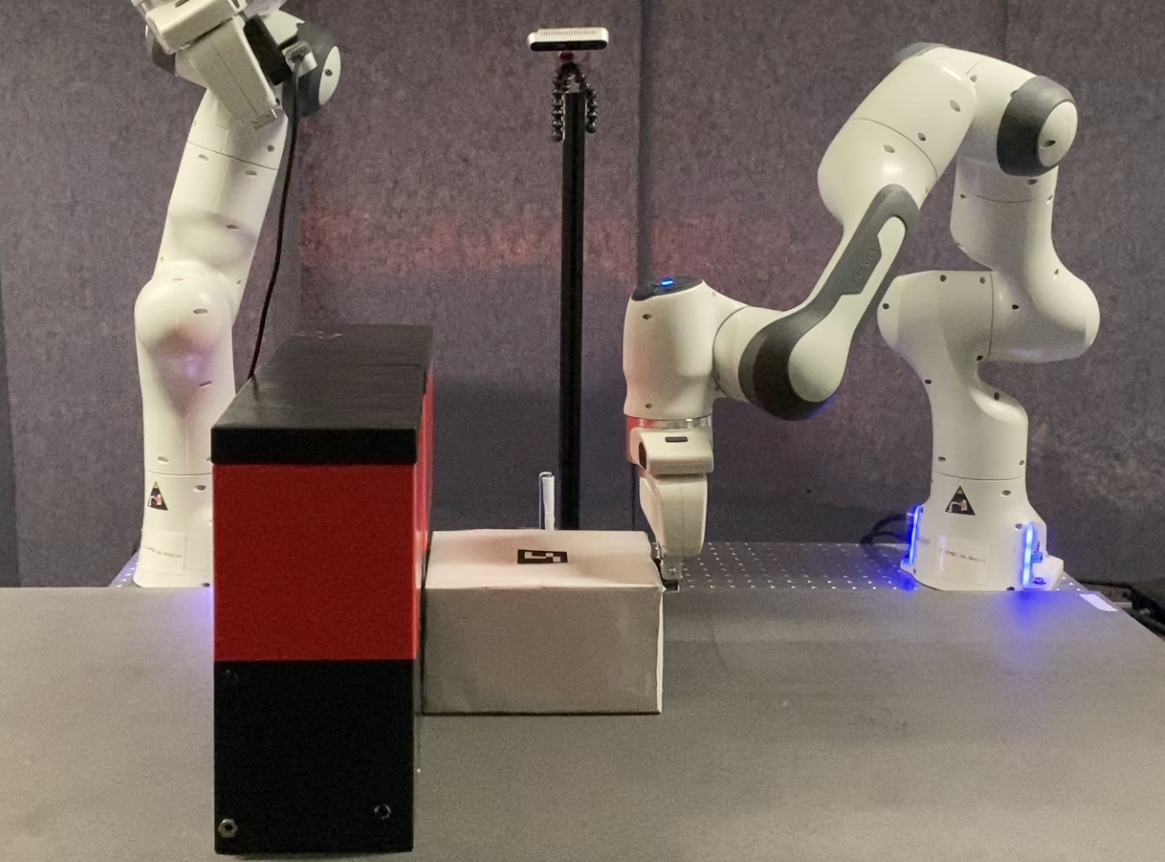}}} \\
	\subfloat[Pivoting]{{\includegraphics[width=0.5\columnwidth]{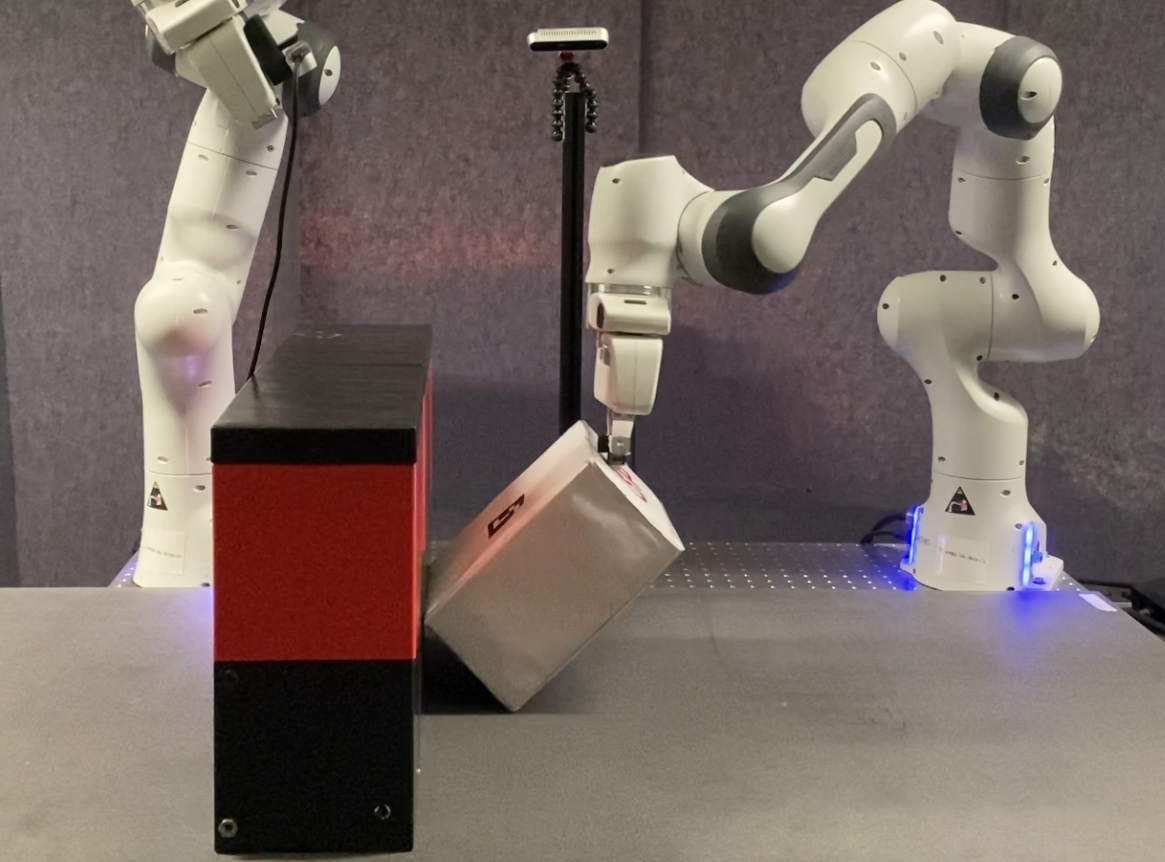}}} 
	\subfloat[Pulling]{{\includegraphics[width=0.5\columnwidth]{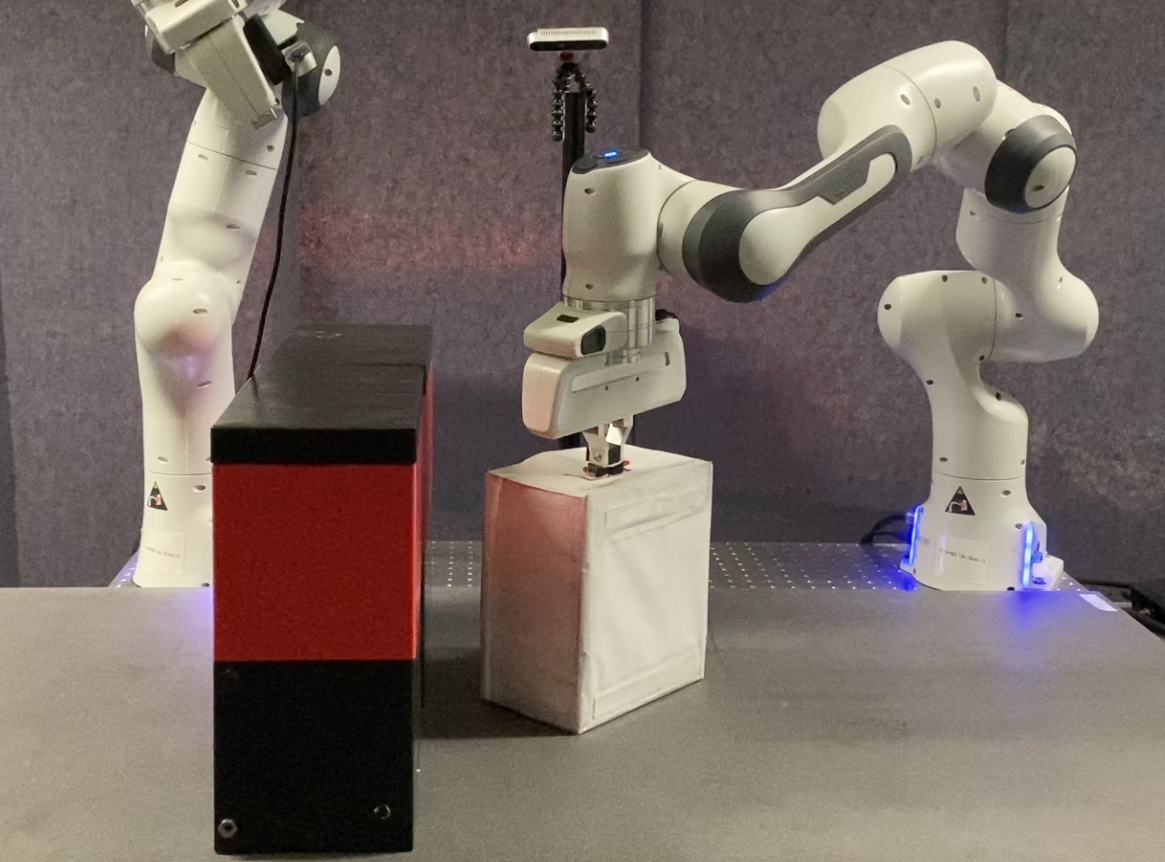}}} 
	\subfloat[Target Reaching]{{\includegraphics[width=0.5\columnwidth]{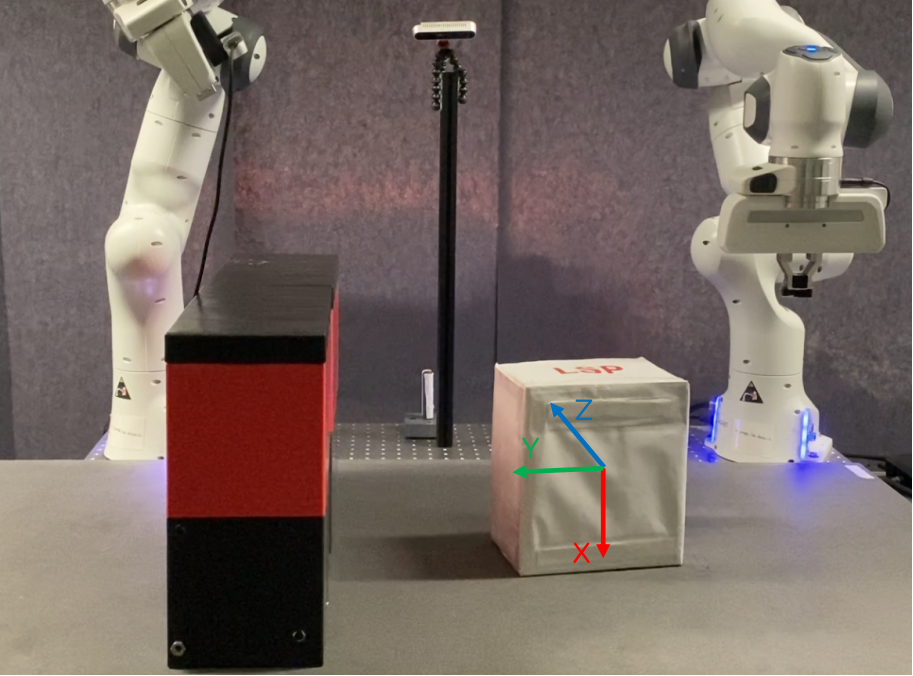}}\label{fig:robot_end}}
	\caption{Non-prehensile manipulation domain task. The system is initialized as (a), and the objective is to manipulate the box to achieve the target configuration as (g). The first stage involves pushing the box towards the wall with a $90^\circ$ rotation. Additionally, we apply an external disturbance to test the skill policy (c). After the pushing stage, the robot switches to the pivoting skill (d, e), followed by pulling (f), until reaching the final geometric configuration.}
	\label{fig:keyframe}
	\vspace{-0.5cm}
\end{figure*}

We conducted real-robot experiments in the Non-Prehensile Manipulation domain, employing a 7-axis Franka Emika robot and a RealSense D435 camera. A large box (21cm x 21cm x 16cm) was positioned on a flat plywood surface. The camera tracked the object motion at 30 Hz with ArUco markers, and the policy for each skill was updated at 100 Hz, with low-level controllers (1000 HZ) actuating the robot. \minew{We employed different controllers depending on the skill operator. Specifically, for pushing, we utilized the Cartesian velocity controller. For other skills, we utilized Cartesian impedance controller. These components are integrated into the Robot Operating System (ROS), enabling their operation at varying frequencies.}

Fig. \ref{fig:keyframe} displays the keyframes of the robot experiments. Given the initial and final configurations, the robot adeptly manipulated the box by utilizing three planar manipulation primitives, establishing and breaking contact with the surroundings. It is worth noting that real-world contact-rich manipulation is quite challenging due to friction uncertainty and external disturbances \cite{pang2023global}. This highlights the importance of introducing skills for online real-time control. Through skill sequencing, we demonstrate that the robot can actively engage with the physical world, accomplishing a much more complex long-horizon task. Additional results are presented in our accompanying video.

\section{Conclusion and Future Work} 
\label{sec:conclusion}

In this paper, we introduced an optimization-based approach for sequential skill planning, namely Logic-Skill Programming (LSP). A first-order extension of a mathematical program is formulated to optimize the overall cumulative reward and the performance of the final configuration. The cumulative reward is abstracted as the sum of value function space. To address such problems, we employed Tensor Train to approximate the value functions and leveraged alternations between symbolic search and skill value optimization to find the optimal solutions.

We demonstrated that the value functions in TT format provide a better approximation of cumulative reward compared to state-of-the-art RL methods. Furthermore, the proposed LSP framework can generate multiple skill skeletons and their corresponding subgoal sequences, given only an evaluation function of the final geometric configuration. We validated this approach across three manipulation domains, highlighting its robust performance in sequencing both prehensile and non-prehensile manipulation primitives.

In this work, the skill learning method, TTPI, assumes a low-rank structure of the value functions for approximation. This works well for skill domains with low-to-medium dimensionality but struggles with image-based policy learning. To address this issue, combining neural networks with TT decomposition could be an interesting direction.

Moreover, the proposed LSP formulation relies on powerful skills to address uncertainty and disturbances. This assumption might not scale well to complicated scenarios, such as pushing with moving obstacles. To address this, a promising approach could be combining global motion planners, such as Rapidly-Exploring Random Tree (RRT)\cite{lavalle1998rapidly}, with the learned skill policies as local controllers.


Furthermore, while Large Language Models (LLMs) have demonstrated impressive results in task planning for robotics, there is still a gap between high-level task planning and low-level control due to the absence of geometric planning. We believe that our method could address this issue by sequencing multiple task-agnostic policies from a skill library. Given a skill skeleton provided by LLMs, our method could assess its feasibility and subsequently returns the optimal subgoal sequence. This subgoal sequence could then be utilized by the sequenced skill policies to actuate the robot in the real world.

\section*{Acknowledgments}
This work was supported by the China Scholarship Council (grant No.202106230104), and by the SWITCH project (\url{https://switch-project.github.io/}), funded by the Swiss National Science Foundation. We would like to thank Yan Zhang for providing helpful feedback on this manuscript, and Jiacheng Qiu for suggestions about the implementation of RL baselines.

\bibliographystyle{plainnat}
\bibliography{references}

\end{document}